\documentclass[acmtog, screen, nonacm]{acmart}

\usepackage{standalone}
\usepackage{tikz}
\usepackage{graphicx}
\usepackage{subcaption}
\usepackage{amsmath,bm}
\usepackage{booktabs} 
\usepackage{eucal}
\usepackage{breqn}
\usepackage{multirow}
\usepackage{wrapfig}
\usepackage[switch]{lineno}
\usepackage[boxed,ruled,linesnumbered]{algorithm2e} 

\SetCommentSty{mycommfont}
\let\oldnl\nl

\newcommand{\nonl}{\renewcommand{\nl}{\let\nl\oldnl}}
\newlength\savedwidth
\newcommand\whline[1]{\noalign{\global\savedwidth\arrayrulewidth
                               \global\arrayrulewidth #1} %
                      \hline
                      \noalign{\global\arrayrulewidth\savedwidth}}
                      
\usepackage[ruled]{algorithm2e} 

\SetAlFnt{\small}
\SetAlCapFnt{\small}
\SetAlCapNameFnt{\small}
\SetAlCapHSkip{0pt}

\begin{document}
\title{3DGS\textsuperscript{2}: Near Second-order Converging 3D Gaussian Splatting}

\author{Lei Lan}
\orcid{0009-0002-7626-7580}
\affiliation{%
 \institution{University of Utah}
 \city{Salt Lake City}
 \state{Utah}
 \country{USA}}
\email{lanlei.virhum@gmail.com}

\author{Tianjia Shao}
\orcid{0000-0001-5485-3752}
\affiliation{%
 \institution{Zhejiang University}
 \city{Zhejiang}
 \country{China}
}
\email{tjshao@zju.edu.cn}

\author{Zixuan Lu}
\orcid{0000-0003-0067-0242}
\affiliation{%
 \institution{University of Utah}
 \city{Salt Lake City}
 \state{Utah}
 \country{USA}}
\email{birdpeople1984@gmail.com}

\author{Yu Zhang}
\orcid{}
\affiliation{%
 \institution{University of Utah}
 \city{Salt Lake City}
 \state{Utah}
 \country{USA}}
\email{zhan723284893@gmail.com}

\author{Chenfanfu Jiang}
\orcid{0000-0003-3506-0583}
\affiliation{%
 \institution{University of California, Los Angeles}
 \city{Los Angeles}
 \country{USA}}
\email{Chenfanfu.Jiang@gmail.com}

\author{Yin Yang}
\orcid{0000-0001-7645-5931}
\affiliation{%
 \institution{University of Utah}
 \city{Salt Lake City}
 \state{Utah}
 \country{USA}
}
\email{yangzzzy@gmail.com}

\begin{abstract}
3D Gaussian Splatting (3DGS) has emerged as a mainstream solution for novel view synthesis and 3D reconstruction. By explicitly encoding a 3D scene using a collection of Gaussian kernels, 3DGS achieves high-quality rendering with superior efficiency. As a learning-based approach, 3DGS training has been dealt with the standard stochastic gradient descent (SGD) method, which offers at most linear convergence. Consequently, training often requires tens of minutes, even with GPU acceleration. This paper introduces a (near) second-order convergent training algorithm for 3DGS, leveraging its unique properties. Our approach is inspired by two key observations. First, the attributes of a Gaussian kernel contribute independently to the image-space loss, which endorses isolated and local optimization algorithms. We exploit this by splitting the optimization at the level of individual kernel attributes, analytically constructing small-size Newton systems for each parameter group, and efficiently solving these systems on GPU threads. This achieves Newton-like convergence per training image without relying on the global Hessian. Second, kernels exhibit sparse and structured coupling across input images. This property allows us to effectively utilize spatial information to mitigate overshoot during stochastic training. Our method converges an order faster than standard GPU-based 3DGS training, requiring over $10 \times$ fewer iterations while maintaining or surpassing the quality of the compared with the SGD-based 3DGS reconstructions. \end{abstract}

%
%


%
%
\begin{teaserfigure}
\centering
\includegraphics[width=\textwidth]{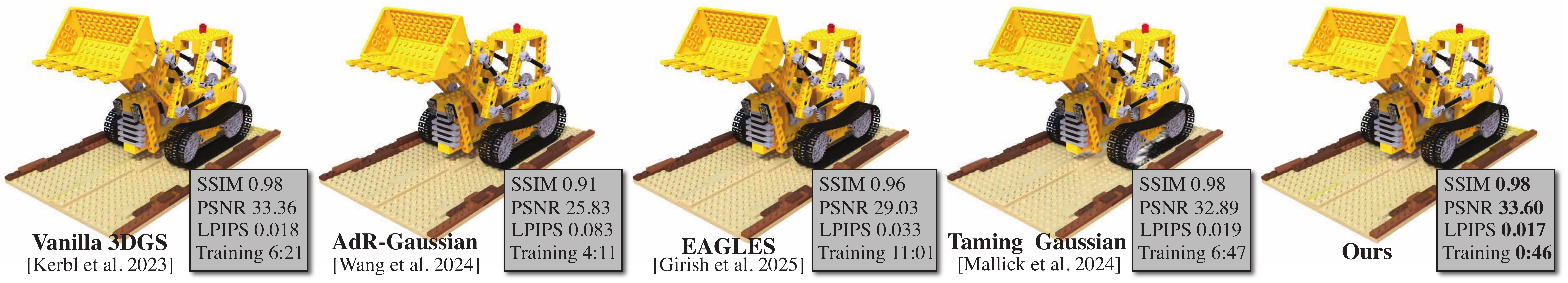}
\captionof{figure}{\textbf{Improve 3DGS training using parallelized local Newton.}~~We show 3DGS training can be substantially improved by switching from stochastic gradient descent to stochastic Newton. A Gaussian kernel has multiple attributes, and optimizing all the attributes of all the kernels as a whole leads to a very high-dimensional nonlinear problem. We leverage the fact that kernel attributes are weakly coupled and design a local Newton scheme to find the optimal value of each type of kernel parameter. We also exploit the spatial relation among input images to effectively mitigate the overshoot issue in stochastic optimization. As a result, our method is $5\times $ to $10\times$ faster than gradient-based 3DGS training. In the teaser, we show the reconstruction results using the-state-of-the-art 3DGS training algorithms, including vanilla 3DGS~\cite{kerbl20233d}, AdR-Gaussian~\cite{adr2024}, EAGLES~\cite{EAGLEs2024}, Taming Gaussian~\cite{Taming3DGS} as well as our method. Our method produces the result of the highest quality (in terms of SSIM, PSNR, and LPIPS) while using only one-tenth of iterations, making the reconstruction of complex 3D scenes in seconds instead of minutes.} 
\label{fig:teaser}
\end{teaserfigure}


\maketitle
\section{Introduction}\label{sec:intro}
The reconstruction of 3D scenes is a fundamental challenge in computer graphics and vision. Recent advancements in learning-based techniques, such as Neural Radiance Fields (NeRFs)~\cite{mildenhall2020nerf}, have re-banded this field. NeRF formulates the reconstruction process as a training task, encoding the color, texture, and geometry of 3D scenes through an implicit MLP network, achieving state-of-the-art results. 
A noteworthy follow-up is 3D Gaussian Splatting (3DGS)~\cite{kerbl20233d}. 3DGS represents a 3D scene explicitly with a collection of Gaussian splatting (GS) kernels. In contrast to NeRF, which relies on ray marching and volume rendering, 3DGS employs rasterization combined with a tile-splatting technique to generate novel views from unseen camera poses. This approach delivers superior runtime efficiency while still maintaining high-quality rendering results.

Just like other learning tasks, stochastic gradient descent (SGD) has been the mainstream modality for training 3DGS~\cite{bottou1998online}. The GS kernels are initialized from SfM (structure-from-motion) \cite{snavely2006photo}, and each kernel is assigned a position, a scaling vector, a rotation, an RGB color, and an opacity. The 3DGS training takes one input image as well as the corresponding camera pose. All the GS kernels are sorted based on the current camera depth, and an image-space loss value is evaluated between the rasterized image and the input image. The loss gradient is then used to update the configurations of all the GS kernels, and the learning rate is empirically adjusted at different stages of the training.

Such an SGD-based first-order optimization has been a default option for many deep-learning tasks, and it works well in general. Nevertheless, we argue that training 3DGS fundamentally differs from generic deep learning. Simply ``copy-and-paste'' the standard learning procedure for 3DGS is neither efficient nor effective. When a neural network is being trained, it is normally believed that parameters housed at neurons are \emph{strongly two-way coupled}, meaning the perturbation at one parameter non-trivially influences all the other parameters as they are connected on the computational graph. This, however, is not the case for 3DGS training, where local greedy choice often suffices. For instance, in order to obtain a good scaling of a GS kernel to suppress the loss, one can always retrieve the best scaling along $x$ direction first, and adjust $y$- and $z$-direction scaling afterwards. In other words, they can be split into multiple sequential local greedy choices. Similarly, the color of a GS kernel does not contribute to a lot of pixels, suggesting color parameters are weakly coupled across GS kernels. Those observations suggest training the GS parameter with the global gradient may be sub-optimal.

In this paper, we propose a novel training algorithm for 3DGS. Our method is nearly second-order convergent based on local Newton's method. Concretely, we prioritize per-kernel parameters and calculate the second-order optimal update for each sub-group of parameters in series. The local updates are executed at all the pertaining GS kernels in parallel in a Jacobi-like manner. Because each local Newton only involves a small number of degrees of freedom (DOFs), we can analytically derive and assemble the local Hessian and gradient at a low cost for each GPU thread. The use of Newton's method avoids the necessity of line search, and the local optimizer always uses the default step size (e.g., the learning rate) of $1.0$. Due to the weak coupling among DOFs, we observe strong second-order convergence during the training. We also deploy a novel sampling strategy, which effectively mitigates the overshoot issue of local optimizer for a given input image. We tested our method on multiple 3DGS datasets, and our algorithm is constantly $10\times$ faster than SGD-based GPU 3DGS training with higher quality, making 3DGS training from minutes to seconds e.g. see Fig.~\ref{fig:teaser}.

\section{Related Work}\label{sec:related}
Novel view synthesis from multi-view images/videos has been a core problem in computer graphics and vision, and a wide range of techniques have been developed for this purpose e.g., see~\cite{NPBG2020,mipnerf2021,hedman2018deep,pnrvo2021,thies2019deferred,lombardi2021mixture,neuralvolume2019,MPI2018,Pulsar2021,Zhang2022sa}. NeRF~\cite{mildenhall2020nerf} implicitly encapsulates the information of the scene with an MLP net. By casting rays over this implicit representation, high-quality scene images can be synthesized. 3DGS~\cite{kerbl20233d}, on the other hand, utilizes a set of GS kernels, which enable more efficient scene rendering via rasterization. They deliver state-of-the-art results and inspire many follow-up research efforts, such as improving the rendering quality~\cite{barron2022mip,Scaffold-GS24,mipgs2024}, increasing the scale of the scene~\cite{kerbl2024hierarchical,song2025city}, 

\subsection{Accelerating NeRF}
 Although NeRF achieves high rendering quality, due to the expensive ray marching and the complexity of MLP, the training and rendering of NeRF are costly. Therefore, many methods have been proposed to accelerate NeRF's training and/or rendering. For example, Sun et al.~\shortcite{DVGO2022} proposed a dense voxel grid storing density and features, along with a post-activation interpolation on voxel density and robust optimization under several priors, leading to very fast convergence of NeRF optimization. Sparse voxel grids with learned features are also utilized to speed up the NeRF rendering~\cite{BakeNerF2021}. KiloNeRF~\cite{KiloNeRF2021} utilizes thousands of small MLPs instead of a single large MLP to reduce the query cost, effectively accelerating the NeRF rendering. The octree data structure can be applied to speed up the rendering~\cite{PlenOctrees2021}. A GPU-friendly implementation~\cite{MobileNeRF2022} also helps NeRF achieve a faster rendering speed. Plenoxels~\cite{Plenoxels2022} replaces MLPs with a sparse 3D grid with spherical harmonics.
The state-of-the-art work is InstantNGP~\cite{InstantNGP2022}, which uses a hash grid and an occupancy grid to accelerate computation and a smaller MLP to represent density and appearance, and its training can be done in a few minutes. While these methods have had great success, due to the ray marching nature, the rendering performance of NeRF-based methods is, in general, slower than that of 3DGS.

\subsection{Accelerating 3DGS}
 3DGS is the state-of-the-art radiance field representation that performs high-quality rendering in real time. Follow-up researches find that its rendering can be further accelerated due to high parameter counts and the redundant or uneven load in rasterization pipeline. 
 For instance, EAGLES~\cite{EAGLEs2024} utilized quantized embeddings to reduce per-point memory storage requirements and a coarse-to-fine training strategy to improve the training convergence speed, which largely reduces storage memory while speeding up the rendering. 
Hamdi et al.~\shortcite{GES2024} introduced Generalized Exponential Functions as a more memory-efficient alternative to Gaussians, which requires fewer particles to represent the scene, reducing the memory storage and increasing the rendering speed.
LightGaussian~\cite{lightGaussian} prunes Gaussians with minimal impact on visual quality, condenses  spherical harmonic coefficients, and applies vector quantization, which substantially compresses Gaussians and largely boosts the rendering speed. C3DGS~\cite{C3DGS24} proposes a learnable mask to remove Gaussians that have minimal impact on overall quality, also reducing the storage and increasing the rendering speed. Speedy-Splat~\cite{hanson2024speedy} precisely localizes Gaussians by computing a tight bounding box around their extent and uses soft and hard pruning to effectively reduce the Gaussian numbers and largely speed up the rendering time as well as training time.
AdR-Gaussian~\cite{adr2024} early culls Gaussian-Tile pairs with low splatting opacity, and proposes a balancing algorithm for pixel thread load, achieving a large increase of rendering and training speed.
Other examples include changing the densification heuristics to reduce the number of Gaussians~\cite{revisitdens2024,minisplat24,3dgsmcmc24} and improving the underlying differentiable rasterizer~\cite{flashgs24,gssplat24}. Their goals are mainly compressing Gaussians and accelerating the Gaussian rendering, with the increased training speed as a byproduct.
Taming 3DGS~\cite{Taming3DGS} utilizes a new densification strategy to reduce Gaussians, together with the backpropagation with per-splat parallelization, to speed up the Gaussian training on limited resources. 3DGS-LM~\cite{3dgslm2024} accelerates the training convergence by replacing the stochastic gradient descent (SGD) optimizer like ADAM~\cite{kingma2014adam} by Levenberg-Marquardt optimizer. The training process converges $30\%$ faster. Nevertheless, the Gauss-Newton optimizer is a pseudo-second-order method. It does not utilize the weak coupling feature of the training. As a result, our method outperforms 3DGS-LM by a significant margin. 
\section{Background}\label{sec:background}
3DGS represents a radiance field using a set of Gaussian kernels $\mathcal{K}$. A GS kernel $k$ is parameterized with a set of trainable attributes, including the center position $\bm{p}_k \in \mathbb{R}^{3}$, the opacity $\sigma_k \in \mathbb{R} $, and the covariance matrix $\bm{A}_k \in \mathbb{R}^{3 \times 3}$. $\bm{A}_k$ is a positive semi-definite definite matrix which only has six independent DOFs. A common practice is to decompose $\bm{A}_k$ as $\bm{A}_k = \bm{R}_k\bm{S}_k\bm{S}_k^\top\bm{R}_k^\top$ with a diagonal scaling matrix $\bm{S}_k = \mathsf{diag}(s_{k,x}, s_{k, y}, s_{k, z})$ and a rotation matrix $\bm{R}_k$. A more compact representation is to use a unit quaternion $\bm{q}_k \in \mathbb{R}^4$, $\|\bm{q}_k\| = 1$ as well as a scaling vector $\bm{s}_k = [s_{k,x}, s_{k, y}, s_{k, z}]^\top$ to encode the orientation and the geometry of the $k$-th Gaussian ellipsoid. A kernel also has its own color information, which is expressed with spherical harmonics (SH) coefficient vector $\bm{c}_k$. To render an image of the scene, GS kernels are sorted based on the current camera depth and projected onto the image plane. The color of an image pixel at $(m, n)$ is computed as:
\begin{equation}\label{eq:3dgs}
\bm{c}(m, n)=\sum_{k \in \mathcal{K}} G_k(m, n) \sigma_k \tilde{\bm{c}}_k(\bm{r}_k, \bm{c}_k) \prod_{j=1}^{k-1}\left(1 - G_j(m, n) \sigma_j \right).
\end{equation}
Here, $G_k(m, n)$ denotes the 2D Gaussian weight of the $k$-th kernel at the pixel whose image coordinate is $(m, n)$. $\bm{r}_k$ stands for the view direction from the camera to the kernel center. $\tilde{\bm{c}}(\bm{r}_k, \bm{c}_k)$ is the view-dependent color information expressed via spherical harmonics (SH). A loss function $L$ measures the difference between the rasterized image and the ground truth image, which consists of a per-pixel color difference and an SSIM loss. 

The training procedure of the vanilla 3DGS searches for a better setup of all the GS kernels by $\bm{x} \leftarrow \alpha \Delta \bm{x}$. Here, $\bm{x}$ is a vector of $|\mathcal{K}|\cdot14$ dimensions, concatenating parameters of all the $| \mathcal{K} |$ kernels. At each iteration, an incremental improvement $\Delta \bm{x}$ is chosen as the negative gradient of the loss $\Delta \bm{x} = -\frac{\partial L}{\partial \bm{x}}$, and the step size $0 < \alpha < 1$ is used to prevent overshooting. The ground-truth loss is defined over all the input images. Because calculating the global gradient is costly, the training uses a sampled gradient based on each input image and progressively reduces the loss in an image-by-image manner.

The above procedure is well-known as a standard way for nonlinear optimization of a high-dimension and nonlinear problem i.e., $\min_{\bm{x}} L(\bm{x})$ using stochastic gradient descent (SGD). SGD is commonly considered linearly convergent as it leverages the first-order Taylor expansion of the target loss~\cite{nocedal1999numerical}. Therefore, it is not surprising to see that 3DGS training needs tens of thousands of iterations to adequately reduce the training loss.

The convergence of the training procedure can be substantially improved using 
Newton's method, which quadratically expands $L(\bm{x}^*)$ at $\bm{x}$ such that:
\begin{equation}\label{eq:newton}
    L(\bm{x}^*) = L^* = L + \bm{g}^\top \Delta \bm{x} + \frac{1}{2} \Delta \bm{x}^\top \bm{H} \Delta \bm{x} + O(\|\Delta \bm{x}\|^3).
\end{equation}
Here, $\bm{x}^*$ is a \emph{local minimizer} of $L$ around current $\bm{x}$. $\bm{g} = \left(\frac{\partial L}{\partial \bm{x}}\right)^\top$, and $\bm{H} = \frac{\partial^2 L}{\partial \bm{x}^2} $ are the gradient and Hessian of the loss $L$. Ignoring the cubic error term, the corresponding DOF update $\Delta \bm{x}$ can be computed from:
\begin{equation*}\label{eq:newton_approximate}
 \Delta \bm{x} = \arg \min_{\bm{y}} L + \bm{g}^\top \bm{y} + \frac{1}{2} \bm{y}^\top \bm{H}\bm{y} 
\end{equation*}
via solving the linear system of:
\begin{equation}\label{eq:newton_system}
    \bm{H}(\bm{x}) \Delta \bm{x} = -\bm{g}(\bm{x}).
\end{equation}
Newton's method possesses many desired properties for nonlinear optimization. First, if $O(\|\Delta \bm{x}\|^3)$ is reasonably small, Newton's method converges quadratically. It normally needs much fewer iterations than GD. More importantly, Newton's method does not need the line search~\cite{nocedal1999numerical}, i.e., $\alpha$ by default is one. When the quadratic approximation of $L^*$ is valid, and $\Delta \bm{x}$ is second-order optimal.

\section{Our Method}\label{sec:method}
Unfortunately, solving the global Newton system at each iteration is computationally expensive, if not impossible. Even the assembly of the matrix is costly and requires tedious implementation. As a result, Newton's method never appears as a viable candidate for 3DGS training. Instead of resorting to the global Newton procedure, we argue that local Newton optimization is also efficient and effective. 
\begin{figure}
    \centering
    \includegraphics[width = 0.8\linewidth]{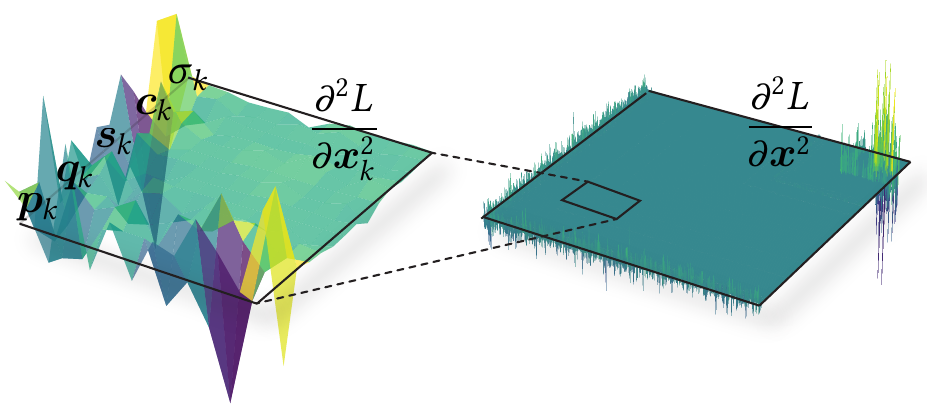}
     \caption{\textbf{Hessian visualization.}~~We plot the values of local and global Hessian matrices for one training image. The local Hessian (left) is for all the parameters of a kernel $\bm{x}_k = [\bm{p}_k^\top, \bm{q}_k^\top, \bm{s}_k^\top, \bm{c}_k^\top, \sigma_k]^\top$, while the global Hessian (right) is for all the kernels' parameters. The variation of the matrices across different DOFs suggests (very) weak coupling amount parameters. }
    \label{fig:hess}
\end{figure}

This reasoning is drawn based on several important observations. First of all, 3DGS training never aims for a global optimum. Just as other deep learning tasks, obsessing over whether a global optimum is achieved holds little practical importance. 
Second, training DOFs at a GS kernel are weakly coupled. Numerically, this can be confirmed by visualizing the full Hessian of  3DGS training. As shown in Fig.~\ref{fig:hess}, there exist sparsely localized ``spikes'' of the values of $\frac{\partial^2 L}{\partial \bm{x}^2}$. This implies the influence among kernel parameters is not global. Solving the full Newton system does not offer a proportional improvement in convergence relative to its computational cost. Similarly, the Hessian of the attributes of one GS kernel is also concentrated, and positional DOFs play a more important role compared with parameters related to color. Departing from generic learning tasks, the training data for 3DGS is highly structured. It is possible for us to exploit the spatial relation among input images to effectively mitigate the oscillation (e.g., overshoot) during the sampled optimization. In the following subsections, we elaborate on the details of the proposed training algorithm.

\subsection{Training data preparation}\label{subsec:training_data_pre}
\setlength{\columnsep}{5 pt}
\begin{wrapfigure}{r}{0.5\linewidth}
    \includegraphics[width=\linewidth]{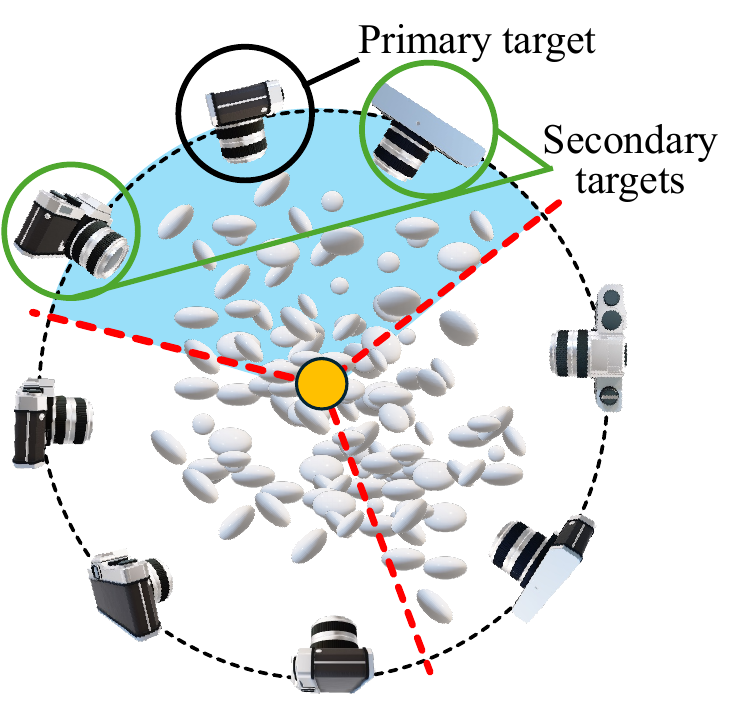}
    \caption{\textbf{Data preparation.}~~We construct a bounding sphere of the scene containing all the GS kernels. Camera poses corresponding to input training images are projected to the sphere surface, and the spherical distance between them is used as the metric to identify KNNs for training images. The current training image is the primary target while its KNNs are the secondary targets.}\label{fig:data_pre}
\end{wrapfigure}
Given a set of input training images $\mathcal{T} = \{I^1, I^2,\cdots, I^{|\mathcal{T}|}\}$, we kick off the training following the vanilla 3DGS by computing a set of initial GS kernels via SfM points. We then estimate a scene center and an encapsulating bounding sphere, which contains all the GS kernels. We project camera poses to the surface of this bounding sphere. The distance between a pair of input images is then defined as the spherical distance between their corresponding camera projections. With the distance defined, we can then obtain the $K$ nearest neighbors (KNNs) of a given input image $I^t$, which intuitively are images that share similar camera perspectives of $I^t$. We call $I^t$ i.e., the current training image, the \emph{primary target}, and the set of its KNNs \emph{secondary targets}, denoted as $\mathcal{N}_t$. Training images that are distant from $I^t$ i.e., $\mathcal{T} - (\{I^t\} + \mathcal{N}_t)$, are considered less relevant as only a small fraction of local kernels are visible from them. On the other hand, secondary targets are more tightly coupled with $I^t$. In our implementation, we set $K = 3$ and noticed this value offers a good balance between convergence and efficiency. As to be discussed later, Hessians and gradients of secondary targets are sparsely evaluated, which accumulate to the primary Hessian/gradient to prevent overshoot (i.e., see Sec.~\ref{subsec:overshoot}). 

\subsection{Stochastic Local Newton}
As mentioned, each GS kernel $k$ houses the position $\bm{p}_k$, opacity $\sigma_k$, orientation $\bm{q}_k$, scaling $\bm{s}_k$, as well as a SH color vector $\bm{c}_k$. Rather than learning the full parameter of $\bm{x}_k = [\bm{p}_k^\top, \sigma_k, \bm{q}_k^\top, \bm{s}_k^\top, \bm{c}_k^\top]^\top$ as one single variable, we split the optimization into multiple sub-steps, following the order of position, orientation, scaling,  opacity, and color. We note that position is of the most importance and should always be optimized first. Shape parameters (orientation, scaling) are prioritized over the color information. This intuition is straightforward to follow: without the right placement of a kernel, the correlated pixels are blank, leaving no opportunities for color-wise adjustment. However, we do not have a strong preference regarding the order of orientation and scaling or color and opacity. Numerically speaking, they become nearly decoupled after the position has been determined.   

Each parameter of kernel $k$ is optimized via a local Newton solve given a certain training input $I^t$:
\begin{equation}\label{eq:local_newton}
    \Delta \bm{y}_k = -\left(\frac{\partial^2 L^t}{\partial \bm{y}_k^2}\right)^{-1} \left(\frac{\partial L^t}{\partial \bm{y}_k}\right)^\top,
\end{equation}
where $\bm{y}_k$ refers to a specific set of variables to be optimized. The loss function depends on the RGB values at a pixel $(m, n)$ from 3DGS rasterization $\bm{c}(m, n)$ and from the training input $I^t$ i.e., $\bm{c}^t(m, n)$. This allows us to use the chain rule to compute the Hessian and gradient of the loss:
\begin{align}\label{eq:local_hessian_gradient}
    \frac{\partial L^t}{\partial \bm{y}_k} = \frac{\partial L^t}{\partial \bm{c}}  \frac{\partial \bm{c}}{\partial \bm{y}_k}, \quad
    \frac{\partial^2 L^t}{\partial \bm{y}_k^2} = \left( \frac{\partial \bm{c}}{\partial \bm{y}_k}\right)^\top \frac{\partial^2 L^t}{\partial \bm{c}^2} \frac{\partial \bm{c}}{\partial \bm{y}_k} + \frac{\partial L^t}{\partial \bm{c}} \cdot \frac{\partial^2 \bm{c}}{\partial \bm{y}^2_k}.
\end{align}

In our implementation, the loss function is the superposition of the L2 loss and SSIM loss such that $L^t = L^t_2 + \lambda L^t_{S}$. To assemble Eq.~\eqref{eq:local_newton} for all the kernel attributes, we first compute the first and second derivatives of L2 loss and SSIM loss w.r.t. the rasterized color:
\begin{align}
& \frac{\partial L^t_2}{\partial \bm{c}(m, n)} = \frac{1}{3|I^t|}\left(\bm{c}(m, n) - \bm{c}^t(m, n)\right), \quad \frac{\partial^2 L^t_2}{\partial \bm{c}^2(m, n)} = \frac{\bm{I}_3}{3|I^t|},\\
& \frac{\partial L^t_S}{\partial \bm{c}(m, n)} = \frac{1}{3|I^t|}\sum\limits_{j=0}^{H}\sum\limits_{i=0}^{W} \frac{\partial f_{S}\left(m + i -\frac{W}{2}, n + j - \frac{H}{2}\right)}{\partial \bm{c}(m, n)},\\
& \frac{\partial^2 L^t_S}{\partial \bm{c}^2(m, n)} =\frac{1}{3|I^t|}\sum\limits_{j=0}^{H}\sum\limits_{i=0}^{W} \frac{\partial^2  f_S\left(m + i - \frac{W}{2}, n + j- \frac{H}{2}\right)}{\partial \bm{c}^2(m, n)}.
\end{align}\\
Here, $|I^t|$ denotes the total number of pixels in the input training $I^t$. $\bm{I}_3$ is a $3 \times 3$ identity matrix. $F_S(a, b)$ measures the SSIM error between $\bm{c}(a, b)$ and $\bm{c}^t(a, b)$ at pixel $(a, b)$~\cite{Wang2004} as the sum of errors of RGB components i.e., $F_S(a, b) = F_{S, R} + F_{S, G} + F_{S, B}$. The error of each color component e.g., the red component $F_{S, R}$ is: 
\begin{equation}
F_{S, R}(a, b) = \frac{(2\mu \mu^t + C_1) (2\tilde{\rho}^t + C_2)}
{\left((\mu \mu^t)^2 + C_1\right) \left((\rho \rho^t)^2 + C_2\right)},
\end{equation}
where $\mu$, $\mu^t$ and $\rho$, $\rho^t$ are respectively the weighted means and variances of the red values within a $W \times H$ Gaussian filter centered at pixel $(a, b)$ in the rasterized image and training image $I^t$. $\tilde{\rho}^t$ is the weighted covariance. $C_1 = 0.01^2 $ and $C_2 = 0.03^2$ are two constants. SSIM uses a filter of $W = H = 11$. $F_{S, G}$ and $F_{S, B}$ are calculated similarly. 

It is clear that $F_{S, R}$ only depends on the red color value, and it is irrelevant to the green or blue components. Therefore, we have:
\begin{align}
    &\frac{\partial F_S(a, b)}{\partial \bm{c}(m, n)} = \frac{\partial F_S}{\partial [c_R, c_G, c_B]^\top} = \left[\frac{\partial F_{S, R}}{c_R}, \frac{\partial F_{S, G}}{c_G}, \frac{\partial F_{S, B}}{c_B}\right],\\
    &\frac{\partial^2 F_S(a, b)}{\partial \bm{c}^2(m, n)} = \mathsf{diag}\left(\frac{\partial^2 F_{S, R}}{c^2_R}, \frac{\partial^2 F_{S, G}}{c^2_G}, \frac{\partial^2 F_{S, B}}{c^2_B}\right).
\end{align}

The first and second derivatives of the red SSIM error w.r.t. the red component (and other color components) can be computed as:
\begin{align}
&\frac{\partial F_{S, R}}{\partial c_R} =\frac{f_1}{f_2f_3} g_0 + \frac{f_0}{f_2f_3} g_1 - \frac{f_0f_1}{f^2_2f_3} g_2 - \frac{f_0f_1}{f_2f^2_3} g_3, \\
&\frac{\partial^2 F_{S, R}}{\partial^2 c_R} = \left(\frac{g_1}{f_2f_3} - \frac{f_2 g_3 + f_3 g_2}{f^2_2f^2_3}f_1\right) g^2_0\nonumber + \left(\frac{g_0}{f_2 f_3} - \frac{f_2 g_3 + f_3 g_2}{f^2_2f^2_3}f_0 \right) g^2_1\nonumber \\
&+\left(\frac{-(f_0 g_1 + f_1 g_0)}{f^2_2f_3} + \frac{2f_2f_3 g_2 +f^2_2 g_3}{(f^2_2f_3)^2}f_0f_1\right) g^2_2\nonumber \\
&+\left(\frac{-(f_0 g_1 + f_1 g_0)}{f_2f^2_3} + \frac{2f_2f_3 g_3 +f^2_3 g_2}{(f_2f^2_3)^2}f_0f_1\right) g^2_3,
\end{align}
where
\begin{align*}
&f_0 = 2\mu \mu^t + C_1, \quad 
f_1 = 2 \tilde{\rho}^t + C_2, \quad
f_3 = (\rho \rho^t)^2 + C_2,\\
& g_0 = 2w_{i,j} \mu^t, \quad g_1 = 2w_{i,j}(\bm{c}(m,n) - \mu^t), \\
& g_2 = 2w_{i,j}\mu, \quad g_3 = 2w_{i,j}(\bm{c}(m,n) - \mu).
\end{align*}
Here, $w_{i,j} = w(m - a+\frac{H}{2}, n - b + \frac{W}{2})$ is the filter's weight.
The per-attribute Hessian and gradient can then be computed with Eq.~\eqref{eq:local_hessian_gradient}, where the first and second derivatives of $\bm{c}(m, n)$ i.e., how the pixel colors change w.r.t. kernel attributes are evaluated at each individual attribute.

\paragraph{Position solve.}
The position of the kernel $k$ is not full-rank. During the optimization, we only permit planner position adjustment of $k$ that is perpendicular to the camera's look at $\bm{r}_k$. To extra the actual DOFs of the kernel position, we build a basis matrix $\bm{U}_k = [\bm{u}_x, \bm{u}_y] \in \mathbb{R}^{3 \times 2}$ such that:
\begin{equation}
    \bm{u}_y = \frac{1}{\bm{r}_k \cdot \bm{r}_k}(\bm{r}_k \otimes \bm{r}_k)[0, 1, 0]^\top,\quad \bm{u}_x = \frac{1}{\sqrt{\bm{r}_k \cdot \bm{r}_k}}\bm{r}_k \times \bm{u}_y.
\end{equation}
The update of the kernel position is then re-parameterized with $\bm{v}_k \in \mathbb{R}^2$:
\begin{equation}
    \Delta \bm{p}_k = \bm{U}_k \Delta \bm{v}_k.
\end{equation}
We first compute the partial derivatives w.r.t. the full positional DOFs via:
\begin{align}
    & \frac{\partial \bm{c}}{\partial \bm{p}_k} =  \frac{\partial\bm{c}}{\partial \tilde{\bm{c}}_k}\frac{\partial\tilde{\bm{c}}_k}{\partial\bm{p}_k} +\frac{\partial \bm{c}}{\partial G_k}\left(\frac{\partial G_k}{\partial \bm{\pi}_k} \frac{\partial \bm{\pi}_k}{\partial \bm{p}_k} +\frac{\partial G_k}{\partial \bm{\Sigma}_k} \frac{\partial \bm{\Sigma}_k}{\partial \bm{p}_k}\right),\\
    & \frac{\partial^2 \bm{c}}{\partial \bm{p}^2_k} = \frac{\partial\bm{c}}{\partial \tilde{\bm{c}}_k} \cdot \frac{\partial^2\tilde{\bm{c}}_k}{\partial\bm{p}^2_k} +
    \frac{\partial \bm{c}}{\partial G_k} 
    \left( 
    \frac{\partial G_k}{\partial \bm{\pi}_k} \cdot \frac{\partial^2 \bm{\pi}_k}{\partial^2 \bm{p}_k} + \frac{\partial G_k}{\partial \Sigma_k} : \frac{\partial^2 \bm{\Sigma}_k}{\partial \bm{p}^2_k} \right. \nonumber \\
    & + \left(\frac{\partial\bm{\pi}_k}{\partial \bm{p}_k}\right)^\top \frac{\partial^2 G_k}{\partial \bm{\pi}^2_k} \frac{\partial\bm{\pi}_k}{\partial \bm{p}_k}  +  \left(\frac{\partial \bm{\Sigma}_k}{\partial \bm{p}_k}\right)^\top : \frac{\partial^2 G_k}{\partial \bm{\Sigma}^2_k} : \frac{\partial \bm{\Sigma}_k}{\partial \bm{p}_k} \nonumber\\
    & \left. + 2 \left(\frac{\partial \bm{\Sigma}_k}{\partial \bm{p}_k}\right)^\top : \frac{\partial^2 G_k}{\partial \bm{\pi}_k\partial \bm{\Sigma}_k} : \frac{\partial \bm{\pi}_k}{\partial \bm{p}_k}\right).
\end{align}
The Hessian and the gradient of the generalized kernel position $\bm{v}_k$ can then be computed as:
\begin{equation}
    \frac{\partial \bm{c}}{\partial \bm{v}_k} = \bm{U}_k^\top \frac{\partial \bm{c}}{\partial \bm{v}_k}, \quad \frac{\partial^2 \bm{c}}{\partial \bm{v}^2_k} = \bm{U}_k^\top \frac{\partial^2 \bm{c}}{\partial \bm{p}^2_k} \bm{U}_k.
\end{equation}
Here, $\bm{\pi}_k = \bm{\pi}_k(\bm{p}_k) \in\mathbb{R}^2$ is the image space projection of the kernel center $\bm{p}_k$ a.k.a. the normalized device coordinate. $\bm{\Sigma}_{k} \in \mathbb{R}^{2 \times 2}$ is 2D covariance matrix. $\tilde{\bm{c}}_k = \tilde{\bm{c}}_k(\bm{r}_k)$ is view-dependent SH color. The first and second derivatives of those intermediate variables are given in the supplementary document.

\paragraph{Scaling solve.}
A potential risk in optimizing scaling and orientation is redundancy. It is the projected 2D Gaussian ellipse that directly influences the pixel color on the rasterized image. Different combinations of scaling and orientation can result in the same projection. This brings the nullspace to the Hessian and undermines the convergence. To avoid this issue, we follow the strategy used for position solve, and perform the local optimization in a reduced space. 

Specifically, to optimize $\bm{s}_k$, we only concern how it would affect the eigenvalue of the 2D covariance matrix $\bm{\Sigma}_{k}$ i.e., the lengths of two axes of the projected ellipse. By eigenvalue decomposition $\bm{\Sigma}_k=\bm{V}_k^\top \bm{\Lambda}_k \bm{V}_k$, we have: 
\begin{equation}
    \bm{\Lambda}_k = \left(\bm{V}_k\bm{J}_k\bm{W}_k\bm{R}_k\right)
    \mathsf{diag} \left(s_{k, x}^2, s_{k, y}^2, s_{k, z}^2\right) \left(\bm{J}_k\bm{W}_k\bm{R}_k\bm{V}_k\right)^\top,
\end{equation}
where $\bm{\Lambda}_k \in\mathbb{R}^{2 \times 2} = \mathsf{diag}(\lambda_{min}, \lambda_{max})$ is the diagonal matrix of eigenvalues. $\bm{V}_k \in\mathbb{R}^{2 \times 3}$ includes the corresponding eigenvectors. $\bm{W}_k$ and $\bm{J}_k$ represent the view matrix and the Jacobian matrix of orthographic projection $\bm{P}$. Since the updates of $s_{k, x}$, $s_{k, y}$, and $s_{k, z}$ only alter $\bm{\Lambda}_k$, $\bm{V}_k$ remains unchanged. $\bm{J}_k$, $\bm{W}_k$ and $\bm{R}_k$ are also constant. Hence, we have:
\begin{equation}
    \left(\bm{V}_k\bm{J}_k\bm{W}_k\bm{R}_k\right)
    \mathsf{diag} \left(s_{k, x}, s_{k, y}, s_{k, z}\right) = \mathsf{diag}(\lambda_{min}, \lambda_{max}),
\end{equation}
leading to
\begin{equation}
    \underbrace{\mathbf{E}_2^\top : \left(\bm{V}_k\bm{J}_k\bm{W}_k\bm{R}_k\right) : \mathbf{E}_3}_{\bm{T}_k} \bm{s}_k = \bm{\lambda}_k = 
    \left[
    \begin{array}{c}
    \lambda_{min}\\
    \lambda_{max}
    \end{array}
    \right],
\end{equation}
where $\mathbf{E}_2$ and $\mathbf{E}_3$ are 3-order tensors padding a vector to a diagonal matrix such that $\bm{\Lambda}_k = \mathbf{E}_2 \cdot \bm{\lambda}_k$ and $\bm{S}_k = \mathbf{E}_3 \cdot \bm{s}_k$. 

The gradient and Hessian of the full scaling factor $\bm{s}_k$ are:
\begin{align}
    &\frac{\partial \bm{c}}{\partial \bm{s}_k} = \frac{\partial \bm{c}}{\partial G_{k}} \frac{\partial G_k}{\partial \bm{\Sigma}_k} : \frac{\partial \bm{\Sigma}_k}{\partial \bm{\lambda}_k} \frac{\partial \bm{\lambda}_k}{\partial \bm{s}_k}, \nonumber\\
    &\frac{\partial^2 \bm{c}}{\partial \bm{s}_k^2} = \frac{\partial \bm{c}}{\partial G_k}\left(\left(\frac{\partial \bm{\Sigma}_k}{\partial \bm{\lambda}_k} \frac{\partial \bm{\lambda}_k}{\partial \bm{s}_k} \right)^\top \frac{\partial^2 G_k}{\partial \bm{\Sigma}^2_k} : \frac{\partial \bm{\Sigma}_k}{\partial \bm{\lambda}_k} \frac{\partial \bm{\lambda}_k}{\partial \bm{s}_k} +\frac{\partial G_k}{\partial \bm{\Sigma}_k} : \frac{\partial \bm{\Sigma}_k}{\partial \bm{\lambda}_k}\frac{\partial^2 \bm{\lambda}_k}{\partial \bm{s}_k^2}\right).
\end{align}
They are then projected into the least-square subspace of $\bm{T}_k$ as:
\begin{equation}
    \frac{\partial \bm{c}}{\partial \bm{\lambda}_k} = \bm{T}_k (\bm{T}_k^\top \bm{T}_k)^{-\top} \frac{\partial \bm{c}}{\partial \bm{s}_k}, \,
    \frac{\partial^2 \bm{c}}{\partial \bm{\lambda}_k^2} = \bm{T}_k (\bm{T}_k^\top \bm{T}_k)^{-\top} \frac{\partial^2 \bm{c}}{\partial \bm{s}_k^2} (\bm{T}_k^\top \bm{T}_k)^{-1}\bm{T}_k^\top. 
\end{equation}

\paragraph{Rotation solve.}
We optimize the rotation $\bm{q}_k$ by constraining its axis to align with $\bm{r}_k$ so the rotation is orthogonal to the scaling. This constraint allows us to re-write the rotation update as $\Delta{\bm{q}_k} = [\cos{\theta_k}, \cos{\theta_k}\bm{r}_k]^\top$, where $\theta_k$ is the rotation angle to be optimized: 
\begin{align}
    &\frac{\partial \bm{c}}{\partial \theta_k}  =\frac{\partial \bm{c}}{\partial G_k}\frac{\partial G_{k}}{\partial \bm{\Sigma}_k}\frac{\partial \bm{\Sigma}_k}{\partial \theta_k},  \nonumber\\
    &\frac{\partial^2 \bm{c}}{\partial \theta_k^2} =\frac{\partial \bm{c}}{\partial G_{k}}\left(\frac{\partial \bm{\Sigma}_k}{\partial\theta_k}^\top \frac{\partial^2 G_{k}}{\partial \bm{\Sigma}^2_k} : \frac{\partial \bm{\Sigma}_k}{\partial \theta_k} + \frac{\partial G_k}{\partial \bm{\Sigma}_k} : \frac{\partial^2 \bm{\Sigma}_k}{\partial \theta^2_k}\right) .
\end{align}
Instead of the linear update $\bm{q} \leftarrow \bm{q} + \Delta \bm{q}$, the rotation update is done with:
\begin{equation}
    \bm{q} \leftarrow \Delta \bm{q} \cdot \bm{q}.
\end{equation}

\paragraph{Opacity solve.}
The opacity $\sigma_k$ is a scalar variable. However, its value is not arbitrary as $\sigma_k \in [0, 1)$, suggesting inequality constraints need to be imposed to avoid out-out-boundary values. We use the interior-point method~\cite{potra2000interior} by adding two logarithmic barrier terms into the local loss such that:
\begin{equation}
    L^t \leftarrow L^t - \alpha_{\sigma} \ln \sigma_k + \ln(1 - \sigma_k).
\end{equation}
The gradient of the opacity is:
\begin{equation}
    \frac{\partial \bm{c}}{\partial \sigma_k} = G_k\left(\prod_{j=1}^{k-1}\left(1 - G_j \sigma_j\right)\right) \left(\bm{c}_k - \sum_{i = k + 1} G_i \sigma_i \bm{c}_i \prod_{j=k+1}^{i-1}\left(1 - G_j \sigma_j\right) \right),    
\end{equation}
and it has a vanished second derivative e.g., $\frac{\partial^2 \bm{c}}{\partial \sigma^2_k} = 0$. The Hessian and gradient w.r.t. $L^t$ should also incorporate barrier loss i.e., $-\frac{1}{\sigma_k} - \frac{1}{ 1 - \sigma_k}$ and $\frac{1}{ (1 - \sigma_k)^2} - \frac{1}{\sigma_k^2}$.

\paragraph{Color solve}
The color of the kernel is encoded with SH bases $\bm{B}_k$, which also depends on the current position of the kernel, i.e., $\bm{B}_k\left(\bm{r}_k(\bm{p}_k)\right)$. Because $\bm{p}_k$ is always the first to be optimized, we decouple this relation during the color solve so that $\bm{r}_k$ is now constant, and $\tilde{\bm{c}}_k = \bm{B}_k \bm{c}_k$ is a linear function of the SH vector $\bm{c}_k$.  

Similar to $F_S$, the optimization of $\bm{c}_k$ can be simplified by de-coupling each color component e.g., between $c_R$ and $\bm{c}_{k, R}$ as:
\begin{align}
    &\frac{\partial c_R}{\partial \bm{c}_{k, R}} =  \sum_{k \in \mathcal{K}} G_k \sigma_k \prod_{j=1}^{k-1}\left(1 - G_j \sigma_j \right) \bm{B}_{k, R},
\end{align}
where $\bm{B}_{k, R}$ is a row vector corresponding to the red component in $\bm{B}_k$. It is easy to see that the second derivative $\frac{\partial^2 c_R}{\partial \bm{c}^2_{k, R}} = 0$ is vanished.


\subsection{Overshoot}\label{subsec:overshoot}
A common challenge for stochastic optimization is \emph{overshoot}. Due to the large amount of training images, it is impractical to evaluate the ground-truth Hessian $\bm{H}^\star$ and gradient $\bm{g}^\star$ because the loss is accumulated over the entire training set $L = L^1 + L^2 + \cdots + L^{|\mathcal{T}|}$.
When we optimize the attributes of GS kernels under the view of the current training image $I^t$, the optimization is not aware of how the result could influence the loss values in other images. If the per-image optimization is over-aggressive, the reduction of $L^t$ can get outweighed by increases at other images, and the global loss $L$ becomes actually worsened. We refer to this numerical phenomenon as overshoot, which severely undermines the global convergence. Preventing overshoot is expensive in general, as the global information of the loss function is needed. As a result, damped batch-based optimization is often employed~\cite{qian1999momentum,kingma2014adam} together with the hyper-parameter tuning of the learning rate. Because of the second-order convergence, overshoot is more likely to occur without the constraint of the learning rate. However, if a learning rate is still necessary, why should we use Newton's method at all? 

We give an efficient solution to this dilemma of stochastic training of 3DGS. Given a training image $I^t$ and the current GS kernel $k$ being trained, we define its \emph{complementary} loss $\bar{L}^t$ such that:
\begin{equation}
    \bar{L}^t = \sum_{j\in\mathcal{T}, j\neq t} L^j,\; \text{for}\; \frac{\partial L^j}{\partial \bm{x}_k} \neq \bm{0}.
\end{equation}
In other words, $\bar{L}^t$ includes all the losses that are relevant to the kernel's attribute. Ideally, if the local training is in the form of $\min_{\bm{x}_k} (L^t + \bar{L}^t)$,
overshoot can be fully avoided. Unfortunately, doing so needs to traverse all the training data, which is clearly prohibitive. Given the spatial arrangement of training images, we note that $\bar{L}^t$ is not evenly distributed over $\mathcal{T}$. Rather, $\bar{L}^t$ concentrates on images sharing similar camera perspectives of $I^t$, i.e., the secondary targets (as mentioned in Sec.~\ref{subsec:training_data_pre}). Therefore, we approximate $\bar{L}^t$ as the total image losses of the secondary targets $\mathcal{N}_t$ and re-form the per-kernel training to the following format:
\begin{equation}\label{eq:tildel}
    \min_{\bm{y}_k} \left( L^t + \tilde{L}^t \right),\; \text{where}\; \tilde{L}^t = \sum_{j \in \mathcal{N}_t} L^j.
\end{equation}

The addition of $\tilde{L}^t$ is intended to prevent overshoot so that $\Delta \bm{y}_k$ does not accidentally increase the loss in $\mathcal{N}_t$. We do not expect $\Delta \bm{y}_k$ to play a primary role in reducing $\tilde{L}^t$ as this is handled by the respective local Newton solve. The actual gradient and Hessian of $\frac{\partial \tilde{L}^t}{\partial \bm{y}_k}$ and $\frac{\partial^2 \tilde{L}^t}{\partial \bm{y}^2_k}$ are of less interest as they act more like a self-adjusting pre-conditioner of $\Delta \bm{y}_k$. Therefore, we downsample images in $\mathcal{N}_t$ for improved training efficiency. The gradient and Hessian of $\tilde{L}^t$ are computed in a similar way as discussed in the previous subsection. 

The experiment is consistent with our analysis (see Sec.~\ref{subsec:secondary_target}). The loss from the secondary targets offers an effective way to mitigate overshoot at a marginal computational cost. Solving Eq.~\eqref{eq:tildel}, the global loss converges stably and hardly shows any oscillations at different training batches without the learning-rate-based step size trimming.

\section{Experimental Results}\label{sec:exp}
We implemented our pipeline on a desktop computer with an \textsf{intel} \textsf{i7-12700} CPU and an \textsf{Nvidia} \textsf{3090} \textsf{RTX} GPU. Our algorithm was implemented using \textsf{C++} and \textsf{CUDA}. We used standard datasets and metrics as in 3DGS~\cite{kerbl20233d}. Specifically, the datasets comprise all scenes from Mip-NeRF360~\cite{barron2022mip}, two scenes from
Tanks \& Temples~\cite{knapitsch2017tanks}, and two scenes from DeepBlending~\cite{hedman2018deep}. Our experiments primarily measure the training convergence, time performance, and the quality of the resulting reconstruction. In general, our training method is $5\times$ to $10\times$ faster than the vanilla GS training using one order fewer iterations. The executable is also available in the supplemental material.

\begin{figure*}
    \includegraphics[width=\linewidth]{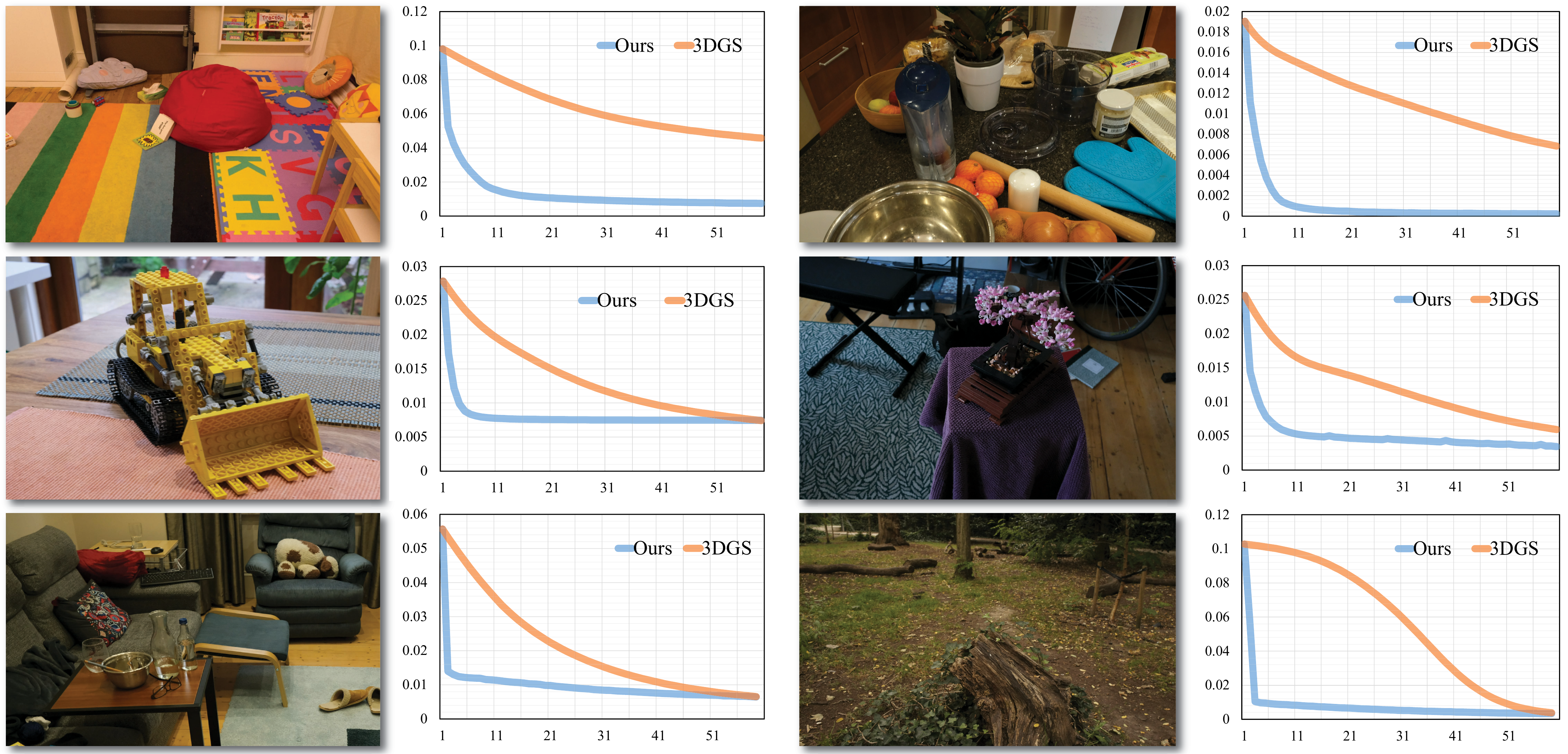}
    \caption{\textbf{Convergence curves of our method and 3DGS.}~~We report a group of representative convergence plots using the proposed training method (local Newton) and vanilla 3DGS training (GD). The corresponding input training images are also attached next to the curves. We observe a strong second-order convergence using our method compared with gradient-based training in almost all the scenes.}\label{fig:vs_gs}
\end{figure*}

A representative example is reported in Fig.~\ref{fig:vs_gs}. In the figure, we plot the convergence curves of 3DGS training of six different scenes given the input image. Being gradient-based, vanilla 3DGS training converges much slower than our method. It is often the case that one iteration using the proposed per-attribute Newton lowers the loss more effectively than $100$ gradient descent iterations. Because local Hessian and gradient are small-size, the computational time for each iteration is only $20\%$ -- $25\%$ slower compared with one gradient descent.

\begin{table*}
\caption{\textbf{Benchmark statistics.}~~This table reports standard benchmarks using different 3DGS training algorithm, including vanilla 3DGS~\cite{kerbl20233d}, AdR-Gaussian~\cite{adr2024}, EAGLES~\cite{EAGLEs2024}, 3DGS-LM~\cite{3dgslm2024}, and Taming 3DGS~\cite{Taming3DGS}. The performance of our method can be further improved with fast rasterization techniques. The table only reports the performance of our method using the standard 3DGS rasterization pipeline.}\label{tab:time}
{\footnotesize \fontfamily{ppl}\selectfont
\begin{center}
\def\arraystretch{1.1}
\begin{tabular}{|c||c|c|c|c|c|c|c|c|c|c|c|c|}
\whline{1.15pt}
\multirow{2}{*}{\textbf{Training algorithm}} & \multicolumn{4}{|c|}{Mip NeRF-360} & \multicolumn{4}{|c|}{Tanks \& Temples} & \multicolumn{4}{|c|}{Deep Blender} \\
\cline{2 - 13}
 & SSIM & PSNR & LPIPS & Training (s) & SSIM & PSNR & LPIPS & Training (s) & SSIM & PSNR & LPIPS & Training (s) \\
\whline{0.5pt}
Vanilla 3DGS~\cite{kerbl20233d} & 0.871 & 29.18 & 0.183 & 1307 & 0.853 & 23.71 & 0.169 & 695 & 0.907 & 29.90 & 0.238 & 1210  \\
AdR-Gaussian~\cite{adr2024} & 0.850 & 28.50 & 0.220 & 783 & 0.835 & 23.52 & 0.201 & 476 & 0.905 & 29.75 & 0.250 & 709  \\
EAGLES~\cite{EAGLEs2024} & 0.810 & 28.64 & 0.192 & 2163 & 0.834 & 23.13 & 0.204 & 913 & 0.909 & 29.79 & 0.242 & 1560 \\
Taming 3DGS~\cite{Taming3DGS} & 0.878 & 29.48 & 0.171 & 781 & 0.859 & 24.15 & 0.161 & 482 & 0.911 & 30.27 & 0.232 & 627 \\
3DGS-LM~\cite{3dgslm2024} & 0.813 & 27.39 & 0.221 & 972 & 0.845 & 23.72 & 0.182 & 663 & 0.903 & 29.72 & 0.247 & 951 \\
\whline{0.5pt}
Our method &0.876 &29.42 &0.168 &$\bm{256}$ ($5.1 \times$)&0.871 &24.43 &0.154 & $\bm{131}$ ($5.3 \times$) &0.927 &30.43 &0.228 & $\bm{189}$ ($6.4 \times$) \\
\whline{1.15pt}
\end{tabular}
\end{center}
}
\end{table*}

\subsection{Optimization order}\label{subsec:order}
\begin{figure*}
    \includegraphics[width=\linewidth]{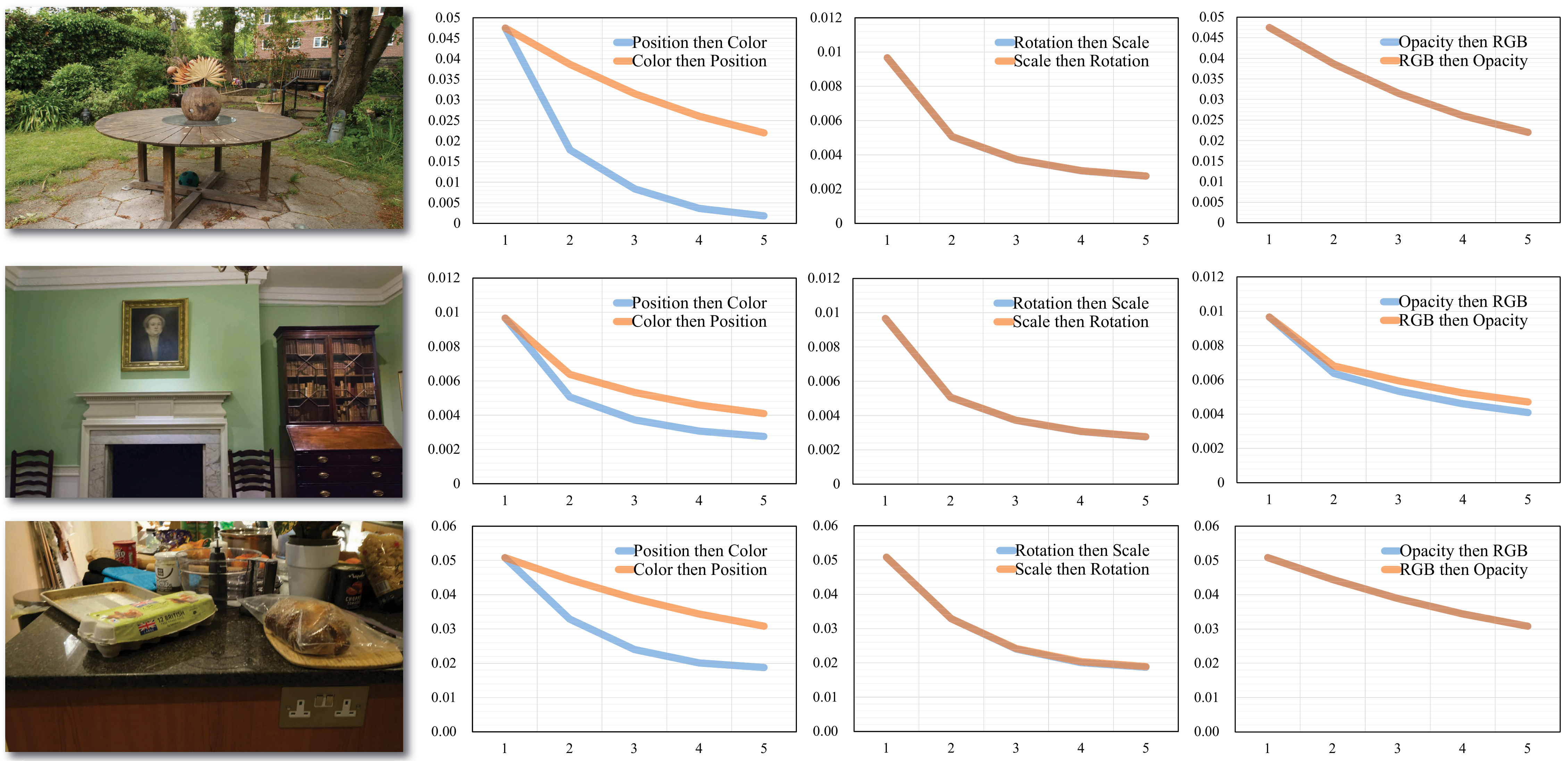}
    \caption{\textbf{Training order of kernel attributes.}~~We always train the kernel position before the color information (RGB and opacity). Without a good positioning of the kernel, there is limited space for color-wise optimization. If one chooses to train the kernel color first, the training converges quadratically as well, but to a different local minimum. Meanwhile, the training order of rotation and scaling or opacity and RBG information is not important. The convergence curves are nearly identical as those attributes are de-coupled.}\label{fig:order}
\end{figure*}

An important strategy allowing local optimization is the decoupling of the kernel attribute. To further validate this, we plot convergence curves of three representative 3DGS reconstruction cases under different orders for per-attribute training. The result is reported in Fig.~\ref{fig:order}. It can be seen that optimizing the positions and geometries of GS kernels should be carried out before color information. Once kernel positions have been determined, orders of scaling and orientation do not matter since they become decoupled. The order between opacity and color is also of less importance. If we choose to optimize the color information first, on the other hand, training also converges but often to a lower-quality local minimum.

\subsection{Trade off of secondary target}\label{subsec:secondary_target}
\begin{figure*}
    \includegraphics[width=0.9 \linewidth]{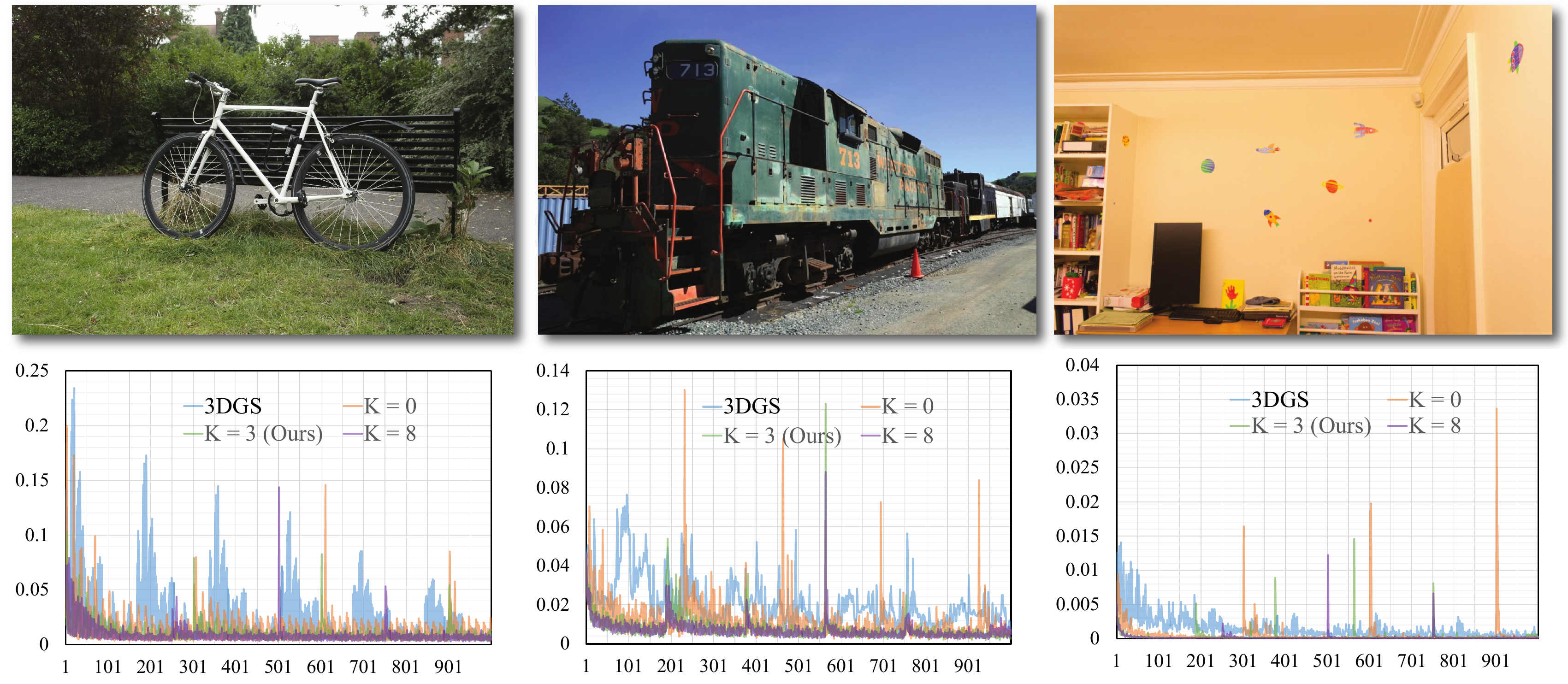}
    \caption{\textbf{Overshoot vs. $|\mathcal{N}_t|$.}~~The Hessian and gradient from secondary target $\mathcal{N}_t$ effectively avoids overshoot, making our training converge smoothly across different batches. The plots show that SGD-based 3DGS training overshoots --- we can see loss spikes after the training switches to a new $I^t$. Without the secondary target i.e., $|\mathcal{N}_t| = 0$, \label{fig:knn}, our method overshoots too, and sometimes even more severely than the vanilla 3DGS, due to its secondary-order nature. Fortunately, sampling at the secondary target of KNNs mitigates this issue. When $K = 8$, we barely see oscillations of the training loss. In our implementation, we use $K = 3$, which balances the training efficiency and convergence.} 
\end{figure*}

Another major contributor to our excellent convergence is mitigating overshoot by estimate $\tilde{L}$ at secondary target $\mathcal{N}_t$. A trade-off we have to decide is the size of $\mathcal{N}_t$ and the convergence rate. Incorporating more training into $\mathcal{N}_t$ improves the convergence but at the cost of more expensive local optimization since the Hessian and gradient of training data in the secondary target are also needed. To this end, we select three classic training scenarios and compare the training time and iteration counts with different $|\mathcal{N}_t|$ i.e., $|\mathcal{N}_t| = 0$, $|\mathcal{N}_t| = 3$, $|\mathcal{N}_t| = 8$. It can be clearly seen from Fig.~\ref{fig:knn} that increasing $|\mathcal{N}_t|$ from $0$ to a small quantity effectively improves the convergence. Further enlarging $\mathcal{N}_t$ to $8$ is able to better resolve this issue at the cost of more expensive computation. This is consistent with our previous analysis and endorses the design of our training algorithm. 

\subsection{Comparison with existing methods}
While many efforts have been investigated to improve the rendering of a reconstructed 3DGS scene, there are relatively fewer works focusing on 3DGS training. We compared our method with vanilla 3DGS~\cite{kerbl20233d}, AdR-Gaussian~\cite{adr2024}, EAGLES~\cite{EAGLEs2024}, 3DGS-LM~\cite{3dgslm2024}, and Taming 3DGS~\cite{Taming3DGS}. The detailed benchmark is reported in Tab.~\ref{tab:time}. AdR-Gaussian~\cite{adr2024} and EAGLES~\cite{EAGLEs2024} focus on improving the efficiency of the rasterization of GS kernels and the memory footprint. Therefore, the total 3DGS training time may also be shortened. Our method is orthogonal to AdR-Gaussian or EAGLES, as we focus on the underlying numerical method for 3DGS training. Combining our method with faster rasterization techniques leads to even faster training. Nevertheless, we report our training performance using the standard 3DGS rasterization pipeline to avoid confusion. 3DGS-LM~\cite{3dgslm2024} replaces SGD with global LM (Levenberg-Marquardt) optimization~\cite{ranganathan2004levenberg} aiming for improved training convergence. Unfortunately, global LM optimization is not able to effectively extract localized nonlinearity (i.e., as shown in Fig.~\ref{fig:hess}). Our method outperforms 3DGS-LM by a significant margin. Taming 3DGS~\cite{Taming3DGS} aims to improve the training efficiency by using fewer GS kernels. It is more effective on hardware platforms with limited computing resources. With our experiment setup, Taming 3DGS performs similarly as AdR-Gaussian. It remains a first-order training modality and is slower than our method.

\section{Conclusion \& Limitation}\label{sec:conclusion}
This paper introduces a second-order convergent training method for 3DGS. We are inspired by the numerical properties of the loss Hessian and switch SGD-based linear optimization to local-Newton-based raining. The decoupled DOFs allow us to analytically derive the local Hessian for each type of kernel attribute and solve the resulting system efficiently in parallel. We also exploit the spatial correlation among input images to approximate the down-sampled global loss so that per-batch 3DGS training does not overshoot. This method outperforms existing method by a significant margin. It demonstrates a strong second-order convergent behavior and only needs one-tenth iterations. As a result, it reduced the total training time by one order without accuracy compromise.

We prove the feasibility of the second-order training for 3DGS reconstruction. While our results are encouraging, there are still many limitations that could be improved in the future. As the vanilla 3DGS, our algorithm is only tested for static scenes. While reconstructing dynamic scenes may be possible, moving GS kernels generates dynamically varying coupling, which may negatively impact the training effectiveness. Evaluating the Hessian of the kernel position is expensive and stands as one major bottleneck of our current pipeline. It may be possible to replace local Newton with Gauss-Newton (e.g. as in~\cite{3dgslm2024} but at local DOFs) or quasi-Newton methods for different kernel attributes to further improve the efficiency. It is also of interest for us to explore auxiliary spatial data structures such as AABB or spatial hashing to speed up the kernel sorting~\cite{lefebvre2006perfect}. 

\bibliographystyle{ACM-Reference-Format}
\bibliography{ref}


\begin{thebibliography}{47}


\ifx \showCODEN    \undefined \def \showCODEN     #1{\unskip}     \fi
\ifx \showDOI      \undefined \def \showDOI       #1{#1}\fi
\ifx \showISBNx    \undefined \def \showISBNx     #1{\unskip}     \fi
\ifx \showISBNxiii \undefined \def \showISBNxiii  #1{\unskip}     \fi
\ifx \showISSN     \undefined \def \showISSN      #1{\unskip}     \fi
\ifx \showLCCN     \undefined \def \showLCCN      #1{\unskip}     \fi
\ifx \shownote     \undefined \def \shownote      #1{#1}          \fi
\ifx \showarticletitle \undefined \def \showarticletitle #1{#1}   \fi
\ifx \showURL      \undefined \def \showURL       {\relax}        \fi
\providecommand\bibfield[2]{#2}
\providecommand\bibinfo[2]{#2}
\providecommand\natexlab[1]{#1}
\providecommand\showeprint[2][]{arXiv:#2}

\bibitem[Aliev et~al\mbox{.}(2020)]%
        {NPBG2020}
\bibfield{author}{\bibinfo{person}{Kara-Ali Aliev}, \bibinfo{person}{Artem Sevastopolsky}, \bibinfo{person}{Maria Kolos}, \bibinfo{person}{Dmitry Ulyanov}, {and} \bibinfo{person}{Victor Lempitsky}.} \bibinfo{year}{2020}\natexlab{}.
\newblock \showarticletitle{Neural Point-Based Graphics}. In \bibinfo{booktitle}{\emph{Computer Vision – ECCV 2020: 16th European Conference, Glasgow, UK, August 23–28, 2020, Proceedings, Part XXII}} (Glasgow, United Kingdom). \bibinfo{publisher}{Springer-Verlag}, \bibinfo{address}{Berlin, Heidelberg}, \bibinfo{pages}{696–712}.
\newblock
\showISBNx{978-3-030-58541-9}


\bibitem[Barron et~al\mbox{.}(2021)]%
        {mipnerf2021}
\bibfield{author}{\bibinfo{person}{Jonathan~T. Barron}, \bibinfo{person}{Ben Mildenhall}, \bibinfo{person}{Matthew Tancik}, \bibinfo{person}{Peter Hedman}, \bibinfo{person}{Ricardo Martin-Brualla}, {and} \bibinfo{person}{Pratul~P. Srinivasan}.} \bibinfo{year}{2021}\natexlab{}.
\newblock \showarticletitle{Mip-NeRF: A Multiscale Representation for Anti-Aliasing Neural Radiance Fields}. In \bibinfo{booktitle}{\emph{2021 IEEE/CVF International Conference on Computer Vision (ICCV)}}. \bibinfo{pages}{5835--5844}.
\newblock


\bibitem[Barron et~al\mbox{.}(2022)]%
        {barron2022mip}
\bibfield{author}{\bibinfo{person}{Jonathan~T Barron}, \bibinfo{person}{Ben Mildenhall}, \bibinfo{person}{Dor Verbin}, \bibinfo{person}{Pratul~P Srinivasan}, {and} \bibinfo{person}{Peter Hedman}.} \bibinfo{year}{2022}\natexlab{}.
\newblock \showarticletitle{Mip-nerf 360: Unbounded anti-aliased neural radiance fields}. In \bibinfo{booktitle}{\emph{Proceedings of the IEEE/CVF conference on computer vision and pattern recognition}}. \bibinfo{pages}{5470--5479}.
\newblock


\bibitem[Bottou(1998)]%
        {bottou1998online}
\bibfield{author}{\bibinfo{person}{L{\'e}on Bottou}.} \bibinfo{year}{1998}\natexlab{}.
\newblock \showarticletitle{Online algorithms and stochastic approximations}.
\newblock \bibinfo{journal}{\emph{Online learning in neural networks}} (\bibinfo{year}{1998}).
\newblock


\bibitem[Chen et~al\mbox{.}(2023)]%
        {MobileNeRF2022}
\bibfield{author}{\bibinfo{person}{Zhiqin Chen}, \bibinfo{person}{Thomas Funkhouser}, \bibinfo{person}{Peter Hedman}, {and} \bibinfo{person}{Andrea Tagliasacchi}.} \bibinfo{year}{2023}\natexlab{}.
\newblock \showarticletitle{MobileNeRF: Exploiting the Polygon Rasterization Pipeline for Efficient Neural Field Rendering on Mobile Architectures}. In \bibinfo{booktitle}{\emph{2023 IEEE/CVF Conference on Computer Vision and Pattern Recognition (CVPR)}}. \bibinfo{pages}{16569--16578}.
\newblock


\bibitem[Fan et~al\mbox{.}(2024)]%
        {lightGaussian}
\bibfield{author}{\bibinfo{person}{Zhiwen Fan}, \bibinfo{person}{Kevin Wang}, \bibinfo{person}{Kairun Wen}, \bibinfo{person}{Zehao Zhu}, \bibinfo{person}{Dejia Xu}, {and} \bibinfo{person}{Zhangyang Wang}.} \bibinfo{year}{2024}\natexlab{}.
\newblock \bibinfo{title}{LightGaussian: Unbounded 3D Gaussian Compression with 15x Reduction and 200+ FPS}.
\newblock
\newblock
\showeprint[arxiv]{2311.17245}~[cs.CV]


\bibitem[Fang and Wang(2024)]%
        {minisplat24}
\bibfield{author}{\bibinfo{person}{Guangchi Fang} {and} \bibinfo{person}{Bing Wang}.} \bibinfo{year}{2024}\natexlab{}.
\newblock \showarticletitle{Mini-Splatting: Representing Scenes with a Constrained Number of Gaussians}. In \bibinfo{booktitle}{\emph{Computer Vision -- ECCV 2024}}, \bibfield{editor}{\bibinfo{person}{Ale{\v{s}} Leonardis}, \bibinfo{person}{Elisa Ricci}, \bibinfo{person}{Stefan Roth}, \bibinfo{person}{Olga Russakovsky}, \bibinfo{person}{Torsten Sattler}, {and} \bibinfo{person}{G{\"u}l Varol}} (Eds.). \bibinfo{publisher}{Springer Nature Switzerland}, \bibinfo{address}{Cham}, \bibinfo{pages}{165--181}.
\newblock
\showISBNx{978-3-031-72980-5}


\bibitem[Feng et~al\mbox{.}(2024)]%
        {flashgs24}
\bibfield{author}{\bibinfo{person}{Guofeng Feng}, \bibinfo{person}{Siyan Chen}, \bibinfo{person}{Rong Fu}, \bibinfo{person}{Zimu Liao}, \bibinfo{person}{Yi Wang}, \bibinfo{person}{Tao Liu}, \bibinfo{person}{Zhilin Pei}, \bibinfo{person}{Hengjie Li}, \bibinfo{person}{Xingcheng Zhang}, {and} \bibinfo{person}{Bo Dai}.} \bibinfo{year}{2024}\natexlab{}.
\newblock \bibinfo{title}{FlashGS: Efficient 3D Gaussian Splatting for Large-scale and High-resolution Rendering}.
\newblock
\newblock
\showeprint[arxiv]{2408.07967}~[cs.CV]


\bibitem[Fridovich-Keil et~al\mbox{.}(2022)]%
        {Plenoxels2022}
\bibfield{author}{\bibinfo{person}{Sara Fridovich-Keil}, \bibinfo{person}{Alex Yu}, \bibinfo{person}{Matthew Tancik}, \bibinfo{person}{Qinhong Chen}, \bibinfo{person}{Benjamin Recht}, {and} \bibinfo{person}{Angjoo Kanazawa}.} \bibinfo{year}{2022}\natexlab{}.
\newblock \showarticletitle{Plenoxels: Radiance Fields without Neural Networks}. In \bibinfo{booktitle}{\emph{2022 IEEE/CVF Conference on Computer Vision and Pattern Recognition (CVPR)}}. \bibinfo{pages}{5491--5500}.
\newblock


\bibitem[Girish et~al\mbox{.}(2025)]%
        {EAGLEs2024}
\bibfield{author}{\bibinfo{person}{Sharath Girish}, \bibinfo{person}{Kamal Gupta}, {and} \bibinfo{person}{Abhinav Shrivastava}.} \bibinfo{year}{2025}\natexlab{}.
\newblock \showarticletitle{EAGLES: Efficient Accelerated 3D Gaussians with Lightweight EncodingS}. In \bibinfo{booktitle}{\emph{Computer Vision -- ECCV 2024}}, \bibfield{editor}{\bibinfo{person}{Ale{\v{s}} Leonardis}, \bibinfo{person}{Elisa Ricci}, \bibinfo{person}{Stefan Roth}, \bibinfo{person}{Olga Russakovsky}, \bibinfo{person}{Torsten Sattler}, {and} \bibinfo{person}{G{\"u}l Varol}} (Eds.). \bibinfo{publisher}{Springer Nature Switzerland}, \bibinfo{address}{Cham}, \bibinfo{pages}{54--71}.
\newblock
\showISBNx{978-3-031-73036-8}


\bibitem[Hamdi et~al\mbox{.}(2024)]%
        {GES2024}
\bibfield{author}{\bibinfo{person}{Abdullah Hamdi}, \bibinfo{person}{Luke Melas-Kyriazi}, \bibinfo{person}{Jinjie Mai}, \bibinfo{person}{Guocheng Qian}, \bibinfo{person}{Ruoshi Liu}, \bibinfo{person}{Carl Vondrick}, \bibinfo{person}{Bernard Ghanem}, {and} \bibinfo{person}{Andrea Vedaldi}.} \bibinfo{year}{2024}\natexlab{}.
\newblock \showarticletitle{GES: Generalized Exponential Splatting for Efficient Radiance Field Rendering}. In \bibinfo{booktitle}{\emph{2024 IEEE/CVF Conference on Computer Vision and Pattern Recognition (CVPR)}}. \bibinfo{pages}{19812--19822}.
\newblock


\bibitem[Hanson et~al\mbox{.}(2024)]%
        {hanson2024speedy}
\bibfield{author}{\bibinfo{person}{Alex Hanson}, \bibinfo{person}{Allen Tu}, \bibinfo{person}{Geng Lin}, \bibinfo{person}{Vasu Singla}, \bibinfo{person}{Matthias Zwicker}, {and} \bibinfo{person}{Tom Goldstein}.} \bibinfo{year}{2024}\natexlab{}.
\newblock \showarticletitle{Speedy-Splat: Fast 3D Gaussian Splatting with Sparse Pixels and Sparse Primitives}.
\newblock \bibinfo{journal}{\emph{arXiv preprint arXiv:2412.00578}} (\bibinfo{year}{2024}).
\newblock


\bibitem[Hedman et~al\mbox{.}(2018)]%
        {hedman2018deep}
\bibfield{author}{\bibinfo{person}{Peter Hedman}, \bibinfo{person}{Julien Philip}, \bibinfo{person}{True Price}, \bibinfo{person}{Jan-Michael Frahm}, \bibinfo{person}{George Drettakis}, {and} \bibinfo{person}{Gabriel Brostow}.} \bibinfo{year}{2018}\natexlab{}.
\newblock \showarticletitle{Deep blending for free-viewpoint image-based rendering}.
\newblock \bibinfo{journal}{\emph{ACM Transactions on Graphics (ToG)}} \bibinfo{volume}{37}, \bibinfo{number}{6} (\bibinfo{year}{2018}), \bibinfo{pages}{1--15}.
\newblock


\bibitem[Hedman et~al\mbox{.}(2021)]%
        {BakeNerF2021}
\bibfield{author}{\bibinfo{person}{Peter Hedman}, \bibinfo{person}{Pratul~P. Srinivasan}, \bibinfo{person}{Ben Mildenhall}, \bibinfo{person}{Jonathan~T. Barron}, {and} \bibinfo{person}{Paul Debevec}.} \bibinfo{year}{2021}\natexlab{}.
\newblock \showarticletitle{Baking Neural Radiance Fields for Real-Time View Synthesis}. In \bibinfo{booktitle}{\emph{2021 IEEE/CVF International Conference on Computer Vision (ICCV)}}. \bibinfo{pages}{5855--5864}.
\newblock


\bibitem[Höllein et~al\mbox{.}(2024)]%
        {3dgslm2024}
\bibfield{author}{\bibinfo{person}{Lukas Höllein}, \bibinfo{person}{Aljaž Božič}, \bibinfo{person}{Michael Zollhöfer}, {and} \bibinfo{person}{Matthias Nießner}.} \bibinfo{year}{2024}\natexlab{}.
\newblock \bibinfo{title}{3DGS-LM: Faster Gaussian-Splatting Optimization with Levenberg-Marquardt}.
\newblock
\newblock
\showeprint[arxiv]{2409.12892}~[cs.CV]


\bibitem[Kerbl et~al\mbox{.}(2023)]%
        {kerbl20233d}
\bibfield{author}{\bibinfo{person}{Bernhard Kerbl}, \bibinfo{person}{Georgios Kopanas}, \bibinfo{person}{Thomas Leimk{\"u}hler}, {and} \bibinfo{person}{George Drettakis}.} \bibinfo{year}{2023}\natexlab{}.
\newblock \showarticletitle{3D Gaussian Splatting for Real-Time Radiance Field Rendering.}
\newblock \bibinfo{journal}{\emph{ACM Trans. Graph.}} \bibinfo{volume}{42}, \bibinfo{number}{4} (\bibinfo{year}{2023}), \bibinfo{pages}{139--1}.
\newblock


\bibitem[Kerbl et~al\mbox{.}(2024)]%
        {kerbl2024hierarchical}
\bibfield{author}{\bibinfo{person}{Bernhard Kerbl}, \bibinfo{person}{Andreas Meuleman}, \bibinfo{person}{Georgios Kopanas}, \bibinfo{person}{Michael Wimmer}, \bibinfo{person}{Alexandre Lanvin}, {and} \bibinfo{person}{George Drettakis}.} \bibinfo{year}{2024}\natexlab{}.
\newblock \showarticletitle{A hierarchical 3d gaussian representation for real-time rendering of very large datasets}.
\newblock \bibinfo{journal}{\emph{ACM Transactions on Graphics (TOG)}} \bibinfo{volume}{43}, \bibinfo{number}{4} (\bibinfo{year}{2024}), \bibinfo{pages}{1--15}.
\newblock


\bibitem[Kheradmand et~al\mbox{.}(2024)]%
        {3dgsmcmc24}
\bibfield{author}{\bibinfo{person}{Shakiba Kheradmand}, \bibinfo{person}{Daniel Rebain}, \bibinfo{person}{Gopal Sharma}, \bibinfo{person}{Weiwei Sun}, \bibinfo{person}{Jeff Tseng}, \bibinfo{person}{Hossam Isack}, \bibinfo{person}{Abhishek Kar}, \bibinfo{person}{Andrea Tagliasacchi}, {and} \bibinfo{person}{Kwang~Moo Yi}.} \bibinfo{year}{2024}\natexlab{}.
\newblock \bibinfo{title}{3D Gaussian Splatting as Markov Chain Monte Carlo}.
\newblock
\newblock
\showeprint[arxiv]{2404.09591}~[cs.CV]


\bibitem[Kingma(2014)]%
        {kingma2014adam}
\bibfield{author}{\bibinfo{person}{Diederik~P Kingma}.} \bibinfo{year}{2014}\natexlab{}.
\newblock \showarticletitle{Adam: A method for stochastic optimization}.
\newblock \bibinfo{journal}{\emph{arXiv preprint arXiv:1412.6980}} (\bibinfo{year}{2014}).
\newblock


\bibitem[Knapitsch et~al\mbox{.}(2017)]%
        {knapitsch2017tanks}
\bibfield{author}{\bibinfo{person}{Arno Knapitsch}, \bibinfo{person}{Jaesik Park}, \bibinfo{person}{Qian-Yi Zhou}, {and} \bibinfo{person}{Vladlen Koltun}.} \bibinfo{year}{2017}\natexlab{}.
\newblock \showarticletitle{Tanks and temples: Benchmarking large-scale scene reconstruction}.
\newblock \bibinfo{journal}{\emph{ACM Transactions on Graphics (ToG)}} \bibinfo{volume}{36}, \bibinfo{number}{4} (\bibinfo{year}{2017}), \bibinfo{pages}{1--13}.
\newblock


\bibitem[Kopanas et~al\mbox{.}(2021)]%
        {pnrvo2021}
\bibfield{author}{\bibinfo{person}{Georgios Kopanas}, \bibinfo{person}{Julien Philip}, \bibinfo{person}{Thomas Leimkühler}, {and} \bibinfo{person}{George Drettakis}.} \bibinfo{year}{2021}\natexlab{}.
\newblock \showarticletitle{Point-Based Neural Rendering with Per-View Optimization}.
\newblock \bibinfo{journal}{\emph{Computer Graphics Forum}} \bibinfo{volume}{40}, \bibinfo{number}{4} (\bibinfo{year}{2021}), \bibinfo{pages}{29--43}.
\newblock


\bibitem[Lassner and Zollhöfer(2021)]%
        {Pulsar2021}
\bibfield{author}{\bibinfo{person}{Christoph Lassner} {and} \bibinfo{person}{Michael Zollhöfer}.} \bibinfo{year}{2021}\natexlab{}.
\newblock \showarticletitle{Pulsar: Efficient Sphere-based Neural Rendering}. In \bibinfo{booktitle}{\emph{2021 IEEE/CVF Conference on Computer Vision and Pattern Recognition (CVPR)}}. \bibinfo{pages}{1440--1449}.
\newblock


\bibitem[Lee et~al\mbox{.}(2024)]%
        {C3DGS24}
\bibfield{author}{\bibinfo{person}{Joo~Chan Lee}, \bibinfo{person}{Daniel Rho}, \bibinfo{person}{Xiangyu Sun}, \bibinfo{person}{Jong~Hwan Ko}, {and} \bibinfo{person}{Eunbyung Park}.} \bibinfo{year}{2024}\natexlab{}.
\newblock \showarticletitle{Compact 3D Gaussian Representation for Radiance Field}. In \bibinfo{booktitle}{\emph{2024 IEEE/CVF Conference on Computer Vision and Pattern Recognition (CVPR)}}. \bibinfo{pages}{21719--21728}.
\newblock


\bibitem[Lefebvre and Hoppe(2006)]%
        {lefebvre2006perfect}
\bibfield{author}{\bibinfo{person}{Sylvain Lefebvre} {and} \bibinfo{person}{Hugues Hoppe}.} \bibinfo{year}{2006}\natexlab{}.
\newblock \showarticletitle{Perfect spatial hashing}.
\newblock \bibinfo{journal}{\emph{ACM Transactions on Graphics (TOG)}} \bibinfo{volume}{25}, \bibinfo{number}{3} (\bibinfo{year}{2006}), \bibinfo{pages}{579--588}.
\newblock


\bibitem[Lombardi et~al\mbox{.}(2019)]%
        {neuralvolume2019}
\bibfield{author}{\bibinfo{person}{Stephen Lombardi}, \bibinfo{person}{Tomas Simon}, \bibinfo{person}{Jason Saragih}, \bibinfo{person}{Gabriel Schwartz}, \bibinfo{person}{Andreas Lehrmann}, {and} \bibinfo{person}{Yaser Sheikh}.} \bibinfo{year}{2019}\natexlab{}.
\newblock \showarticletitle{Neural volumes: learning dynamic renderable volumes from images}.
\newblock \bibinfo{journal}{\emph{ACM Trans. Graph.}} \bibinfo{volume}{38}, \bibinfo{number}{4}, Article \bibinfo{articleno}{65} (\bibinfo{date}{July} \bibinfo{year}{2019}), \bibinfo{numpages}{14}~pages.
\newblock
\showISSN{0730-0301}


\bibitem[Lombardi et~al\mbox{.}(2021)]%
        {lombardi2021mixture}
\bibfield{author}{\bibinfo{person}{Stephen Lombardi}, \bibinfo{person}{Tomas Simon}, \bibinfo{person}{Gabriel Schwartz}, \bibinfo{person}{Michael Zollhoefer}, \bibinfo{person}{Yaser Sheikh}, {and} \bibinfo{person}{Jason Saragih}.} \bibinfo{year}{2021}\natexlab{}.
\newblock \showarticletitle{Mixture of volumetric primitives for efficient neural rendering}.
\newblock \bibinfo{journal}{\emph{ACM Transactions on Graphics (ToG)}} \bibinfo{volume}{40}, \bibinfo{number}{4} (\bibinfo{year}{2021}), \bibinfo{pages}{1--13}.
\newblock


\bibitem[Lu et~al\mbox{.}(2024)]%
        {Scaffold-GS24}
\bibfield{author}{\bibinfo{person}{Tao Lu}, \bibinfo{person}{Mulin Yu}, \bibinfo{person}{Linning Xu}, \bibinfo{person}{Yuanbo Xiangli}, \bibinfo{person}{Limin Wang}, \bibinfo{person}{Dahua Lin}, {and} \bibinfo{person}{Bo Dai}.} \bibinfo{year}{2024}\natexlab{}.
\newblock \showarticletitle{Scaffold-GS: Structured 3D Gaussians for View-Adaptive Rendering}. In \bibinfo{booktitle}{\emph{2024 IEEE/CVF Conference on Computer Vision and Pattern Recognition (CVPR)}}. \bibinfo{pages}{20654--20664}.
\newblock


\bibitem[Mallick et~al\mbox{.}(2024)]%
        {Taming3DGS}
\bibfield{author}{\bibinfo{person}{Saswat~Subhajyoti Mallick}, \bibinfo{person}{Rahul Goel}, \bibinfo{person}{Bernhard Kerbl}, \bibinfo{person}{Markus Steinberger}, \bibinfo{person}{Francisco~Vicente Carrasco}, {and} \bibinfo{person}{Fernando De~La~Torre}.} \bibinfo{year}{2024}\natexlab{}.
\newblock \showarticletitle{Taming 3DGS: High-Quality Radiance Fields with Limited Resources}. In \bibinfo{booktitle}{\emph{SIGGRAPH Asia 2024 Conference Papers}} \emph{(\bibinfo{series}{SA '24})}. \bibinfo{publisher}{Association for Computing Machinery}, \bibinfo{address}{New York, NY, USA}, Article \bibinfo{articleno}{2}, \bibinfo{numpages}{11}~pages.
\newblock
\showISBNx{9798400711312}


\bibitem[Mildenhall et~al\mbox{.}(2020)]%
        {mildenhall2020nerf}
\bibfield{author}{\bibinfo{person}{B Mildenhall}, \bibinfo{person}{PP Srinivasan}, \bibinfo{person}{M Tancik}, \bibinfo{person}{JT Barron}, \bibinfo{person}{R Ramamoorthi}, {and} \bibinfo{person}{R Ng}.} \bibinfo{year}{2020}\natexlab{}.
\newblock \showarticletitle{Nerf: Representing scenes as neural radiance fields for view synthesis}. In \bibinfo{booktitle}{\emph{European conference on computer vision}}.
\newblock


\bibitem[M\"{u}ller et~al\mbox{.}(2022)]%
        {InstantNGP2022}
\bibfield{author}{\bibinfo{person}{Thomas M\"{u}ller}, \bibinfo{person}{Alex Evans}, \bibinfo{person}{Christoph Schied}, {and} \bibinfo{person}{Alexander Keller}.} \bibinfo{year}{2022}\natexlab{}.
\newblock \showarticletitle{Instant neural graphics primitives with a multiresolution hash encoding}.
\newblock  \bibinfo{volume}{41}, \bibinfo{number}{4}, Article \bibinfo{articleno}{102} (\bibinfo{date}{July} \bibinfo{year}{2022}), \bibinfo{numpages}{15}~pages.
\newblock
\showISSN{0730-0301}


\bibitem[Nocedal and Wright(1999)]%
        {nocedal1999numerical}
\bibfield{author}{\bibinfo{person}{Jorge Nocedal} {and} \bibinfo{person}{Stephen~J Wright}.} \bibinfo{year}{1999}\natexlab{}.
\newblock \bibinfo{booktitle}{\emph{Numerical optimization}}.
\newblock \bibinfo{publisher}{Springer}.
\newblock


\bibitem[Potra and Wright(2000)]%
        {potra2000interior}
\bibfield{author}{\bibinfo{person}{Florian~A Potra} {and} \bibinfo{person}{Stephen~J Wright}.} \bibinfo{year}{2000}\natexlab{}.
\newblock \showarticletitle{Interior-point methods}.
\newblock \bibinfo{journal}{\emph{Journal of computational and applied mathematics}} \bibinfo{volume}{124}, \bibinfo{number}{1-2} (\bibinfo{year}{2000}), \bibinfo{pages}{281--302}.
\newblock


\bibitem[Qian(1999)]%
        {qian1999momentum}
\bibfield{author}{\bibinfo{person}{Ning Qian}.} \bibinfo{year}{1999}\natexlab{}.
\newblock \showarticletitle{On the momentum term in gradient descent learning algorithms}.
\newblock \bibinfo{journal}{\emph{Neural networks}} \bibinfo{volume}{12}, \bibinfo{number}{1} (\bibinfo{year}{1999}), \bibinfo{pages}{145--151}.
\newblock


\bibitem[Ranganathan(2004)]%
        {ranganathan2004levenberg}
\bibfield{author}{\bibinfo{person}{Ananth Ranganathan}.} \bibinfo{year}{2004}\natexlab{}.
\newblock \showarticletitle{The levenberg-marquardt algorithm}.
\newblock \bibinfo{journal}{\emph{Tutoral on LM algorithm}} \bibinfo{volume}{11}, \bibinfo{number}{1} (\bibinfo{year}{2004}), \bibinfo{pages}{101--110}.
\newblock


\bibitem[Reiser et~al\mbox{.}(2021)]%
        {KiloNeRF2021}
\bibfield{author}{\bibinfo{person}{Christian Reiser}, \bibinfo{person}{Songyou Peng}, \bibinfo{person}{Yiyi Liao}, {and} \bibinfo{person}{Andreas Geiger}.} \bibinfo{year}{2021}\natexlab{}.
\newblock \showarticletitle{KiloNeRF: Speeding up Neural Radiance Fields with Thousands of Tiny MLPs}. In \bibinfo{booktitle}{\emph{2021 IEEE/CVF International Conference on Computer Vision (ICCV)}}. \bibinfo{pages}{14315--14325}.
\newblock


\bibitem[Rota~Bul{\`o} et~al\mbox{.}(2025)]%
        {revisitdens2024}
\bibfield{author}{\bibinfo{person}{Samuel Rota~Bul{\`o}}, \bibinfo{person}{Lorenzo Porzi}, {and} \bibinfo{person}{Peter Kontschieder}.} \bibinfo{year}{2025}\natexlab{}.
\newblock \showarticletitle{Revising Densification in Gaussian Splatting}. In \bibinfo{booktitle}{\emph{Computer Vision -- ECCV 2024}}, \bibfield{editor}{\bibinfo{person}{Ale{\v{s}} Leonardis}, \bibinfo{person}{Elisa Ricci}, \bibinfo{person}{Stefan Roth}, \bibinfo{person}{Olga Russakovsky}, \bibinfo{person}{Torsten Sattler}, {and} \bibinfo{person}{G{\"u}l Varol}} (Eds.). \bibinfo{publisher}{Springer Nature Switzerland}, \bibinfo{address}{Cham}, \bibinfo{pages}{347--362}.
\newblock
\showISBNx{978-3-031-73036-8}


\bibitem[Snavely et~al\mbox{.}(2006)]%
        {snavely2006photo}
\bibfield{author}{\bibinfo{person}{Noah Snavely}, \bibinfo{person}{Steven~M Seitz}, {and} \bibinfo{person}{Richard Szeliski}.} \bibinfo{year}{2006}\natexlab{}.
\newblock \showarticletitle{Photo tourism: exploring photo collections in 3D}.
\newblock In \bibinfo{booktitle}{\emph{ACM siggraph 2006 papers}}. \bibinfo{pages}{835--846}.
\newblock


\bibitem[Song et~al\mbox{.}(2025)]%
        {song2025city}
\bibfield{author}{\bibinfo{person}{Kaiwen Song}, \bibinfo{person}{Xiaoyi Zeng}, \bibinfo{person}{Chenqu Ren}, {and} \bibinfo{person}{Juyong Zhang}.} \bibinfo{year}{2025}\natexlab{}.
\newblock \showarticletitle{City-on-web: Real-time neural rendering of large-scale scenes on the web}. In \bibinfo{booktitle}{\emph{European Conference on Computer Vision}}. Springer, \bibinfo{pages}{385--402}.
\newblock


\bibitem[Sun et~al\mbox{.}(2022)]%
        {DVGO2022}
\bibfield{author}{\bibinfo{person}{Cheng Sun}, \bibinfo{person}{Min Sun}, {and} \bibinfo{person}{Hwann-Tzong Chen}.} \bibinfo{year}{2022}\natexlab{}.
\newblock \showarticletitle{Direct Voxel Grid Optimization: Super-fast Convergence for Radiance Fields Reconstruction}. In \bibinfo{booktitle}{\emph{2022 IEEE/CVF Conference on Computer Vision and Pattern Recognition (CVPR)}}. \bibinfo{pages}{5449--5459}.
\newblock


\bibitem[Thies et~al\mbox{.}(2019)]%
        {thies2019deferred}
\bibfield{author}{\bibinfo{person}{Justus Thies}, \bibinfo{person}{Michael Zollh{\"o}fer}, {and} \bibinfo{person}{Matthias Nie{\ss}ner}.} \bibinfo{year}{2019}\natexlab{}.
\newblock \showarticletitle{Deferred neural rendering: Image synthesis using neural textures}.
\newblock \bibinfo{journal}{\emph{Acm Transactions on Graphics (TOG)}} \bibinfo{volume}{38}, \bibinfo{number}{4} (\bibinfo{year}{2019}), \bibinfo{pages}{1--12}.
\newblock


\bibitem[Wang et~al\mbox{.}(2024)]%
        {adr2024}
\bibfield{author}{\bibinfo{person}{Xinzhe Wang}, \bibinfo{person}{Ran Yi}, {and} \bibinfo{person}{Lizhuang Ma}.} \bibinfo{year}{2024}\natexlab{}.
\newblock \showarticletitle{AdR-Gaussian: Accelerating Gaussian Splatting with Adaptive Radius}. In \bibinfo{booktitle}{\emph{SIGGRAPH Asia 2024 Conference Papers}} \emph{(\bibinfo{series}{SA '24})}. \bibinfo{publisher}{Association for Computing Machinery}, \bibinfo{address}{New York, NY, USA}, Article \bibinfo{articleno}{73}, \bibinfo{numpages}{10}~pages.
\newblock
\showISBNx{9798400711312}


\bibitem[Wang et~al\mbox{.}(2004)]%
        {Wang2004}
\bibfield{author}{\bibinfo{person}{Zhou Wang}, \bibinfo{person}{A.C. Bovik}, \bibinfo{person}{H.R. Sheikh}, {and} \bibinfo{person}{E.P. Simoncelli}.} \bibinfo{year}{2004}\natexlab{}.
\newblock \showarticletitle{Image quality assessment: from error visibility to structural similarity}.
\newblock \bibinfo{journal}{\emph{IEEE Transactions on Image Processing}} \bibinfo{volume}{13}, \bibinfo{number}{4} (\bibinfo{year}{2004}), \bibinfo{pages}{600--612}.
\newblock


\bibitem[Ye et~al\mbox{.}(2024)]%
        {gssplat24}
\bibfield{author}{\bibinfo{person}{Vickie Ye}, \bibinfo{person}{Ruilong Li}, \bibinfo{person}{Justin Kerr}, \bibinfo{person}{Matias Turkulainen}, \bibinfo{person}{Brent Yi}, \bibinfo{person}{Zhuoyang Pan}, \bibinfo{person}{Otto Seiskari}, \bibinfo{person}{Jianbo Ye}, \bibinfo{person}{Jeffrey Hu}, \bibinfo{person}{Matthew Tancik}, {and} \bibinfo{person}{Angjoo Kanazawa}.} \bibinfo{year}{2024}\natexlab{}.
\newblock \bibinfo{title}{gsplat: An Open-Source Library for Gaussian Splatting}.
\newblock
\newblock
\showeprint[arxiv]{2409.06765}~[cs.CV]


\bibitem[Yu et~al\mbox{.}(2021)]%
        {PlenOctrees2021}
\bibfield{author}{\bibinfo{person}{Alex Yu}, \bibinfo{person}{Ruilong Li}, \bibinfo{person}{Matthew Tancik}, \bibinfo{person}{Hao Li}, \bibinfo{person}{Ren Ng}, {and} \bibinfo{person}{Angjoo Kanazawa}.} \bibinfo{year}{2021}\natexlab{}.
\newblock \showarticletitle{{ PlenOctrees for Real-time Rendering of Neural Radiance Fields }}. In \bibinfo{booktitle}{\emph{2021 IEEE/CVF International Conference on Computer Vision (ICCV)}}. \bibinfo{publisher}{IEEE Computer Society}, \bibinfo{address}{Los Alamitos, CA, USA}, \bibinfo{pages}{5732--5741}.
\newblock


\bibitem[Yu et~al\mbox{.}(2024)]%
        {mipgs2024}
\bibfield{author}{\bibinfo{person}{Zehao Yu}, \bibinfo{person}{Anpei Chen}, \bibinfo{person}{Binbin Huang}, \bibinfo{person}{Torsten Sattler}, {and} \bibinfo{person}{Andreas Geiger}.} \bibinfo{year}{2024}\natexlab{}.
\newblock \showarticletitle{Mip-Splatting: Alias-Free 3D Gaussian Splatting}. In \bibinfo{booktitle}{\emph{2024 IEEE/CVF Conference on Computer Vision and Pattern Recognition (CVPR)}}. \bibinfo{pages}{19447--19456}.
\newblock


\bibitem[Zhang et~al\mbox{.}(2022)]%
        {Zhang2022sa}
\bibfield{author}{\bibinfo{person}{Qiang Zhang}, \bibinfo{person}{Seung-Hwan Baek}, \bibinfo{person}{Szymon Rusinkiewicz}, {and} \bibinfo{person}{Felix Heide}.} \bibinfo{year}{2022}\natexlab{}.
\newblock \showarticletitle{Differentiable Point-Based Radiance Fields for Efficient View Synthesis}. In \bibinfo{booktitle}{\emph{SIGGRAPH Asia 2022 Conference Papers}} (Daegu, Republic of Korea,). \bibinfo{publisher}{ACM}, \bibinfo{address}{New York, NY, USA}, Article \bibinfo{articleno}{7}.
\newblock
\showISBNx{9781450394703}


\bibitem[Zhou et~al\mbox{.}(2018)]%
        {MPI2018}
\bibfield{author}{\bibinfo{person}{Tinghui Zhou}, \bibinfo{person}{Richard Tucker}, \bibinfo{person}{John Flynn}, \bibinfo{person}{Graham Fyffe}, {and} \bibinfo{person}{Noah Snavely}.} \bibinfo{year}{2018}\natexlab{}.
\newblock \showarticletitle{Stereo magnification: learning view synthesis using multiplane images}.
\newblock \bibinfo{journal}{\emph{ACM Trans. Graph.}} \bibinfo{volume}{37}, \bibinfo{number}{4}, Article \bibinfo{articleno}{65} (\bibinfo{date}{July} \bibinfo{year}{2018}), \bibinfo{numpages}{12}~pages.
\newblock
\showISSN{0730-0301}


\end{thebibliography}
\begin{appendix}

\section{Derivatives of Intermediate Variables}
We give the detailed formulation of the first and second derivatives of intermediate variables w.r.t. the kernel center $\bm{p}_k$.

\subsection{Projection function}
$\bm{\pi}_k(\bm{p}_k)$ is the projection function, which converts the kernel center from 3D space to its 2D normalized device coordinate. Its first and second derivative w.r.t. the kernel center is:
\begin{align}
& \frac{\partial\bm{\pi}_k}{\partial\bm{p}_k} = 
\begin{bmatrix}
\frac{W_I^t}{2}\left(\frac{1}{h_w}(\bm{P}\bm{W})_{0} - \frac{h_x}{h_w}(\bm{P}\bm{W})_3\right)^\top\\
\frac{H_I^t}{2}\left(\frac{1}{h_w}(\bm{P}\bm{W})_{1} - \frac{h_y}{h_w}(\bm{P}\bm{W})_{3}\right)^\top
\end{bmatrix},\\
& \frac{\partial^2\bm{\pi}_k}{\partial \bm{p}^2_k} = 
\begin{bmatrix}
W^t_I \left(\frac{h_x}{h^2_w}(\bm{P}\bm{W})^\top_3(\bm{P}\bm{W})_{3} - \frac{1}{h^2_w}(\bm{P}\bm{W})^\top_{3}(\bm{P}\bm{W})_0\right)\\
H^t_I \left(\frac{h_y}{h^2_w}(\bm{P}\bm{W})^\top_{3}(\bm{P}\bm{W})_3 - \frac{1}{h^2_w}(\bm{P}\bm{W})^\top_{3}(\bm{P}\bm{W})_1\right)
\end{bmatrix}.
\end{align}
Here, $W_I^t$ and $H_I^t$ are the width and height of the input training $I^t$. $\bm{P}$ and $\bm{W}$ are $4\times 4$ (homogeneous) projection matrix and viewing matrix. 
\begin{equation}
    \bm{h} = [h_x, h_y, h_z, h_w]^\top= \bm{P} \bm{W} [\bm{p}^\top_k,1]^\top \in \mathbb{R}^4.
\end{equation}
The notion $(\bm{P}\bm{W})_i $ is a 3 dimensional \emph{row vector} corresponding to the first three entries of $i$-th row of $\bm{P}\bm{W}$.


\subsection{SH color}
$\tilde{\bm{c}}_k \in \mathbb{R}^3$ is the view-dependent color based on the SH coefficients. We give the derivatives for the red color component ($\tilde{c}_{k, R}$), and the formulation for the other two components is the same.  
\begin{align}
&\frac{\partial\tilde{c}_{k,R}}{\partial\bm{p}_k} = \left(\bm{B}_{k, R} \odot \bm{c}_{k,R}\right)^\top \frac{\partial \bm{\Phi}(\bm{r}_k)}{\partial \bm{r}_k}:\frac{\partial \bm{\Phi}(\bm{r}_k)}{\partial \bm{p}_k}, \nonumber\\
&\frac{\partial^2\tilde{c}_{k, R}}{\partial\bm{p}^2_k} = \left(\frac{\partial \Phi(\bm{r}_k)}{\partial \bm{p}_k}\right)^\top : \left(\bm{B}_{k, R} \odot \bm{c}_{k,R}\right)^\top \frac{\partial^2 \bm{\Phi}(\bm{r}_k)}{\partial \bm{r}^2_k}:\frac{\partial \bm{\Phi}(\bm{r}_k)}{\partial \bm{p}_k} \nonumber \\
& + \left(\bm{B}_{k, R} \odot \bm{c}_{k,R}\right)^\top \frac{\partial \Phi(\bm{r}_k)}{\partial \bm{r}_k}:\frac{\partial^2 \bm{\Phi}(\bm{r}_k)}{\partial \bm{p}^2_k}.
\end{align}
$\bm{B}_k$ is the SH bases. $\odot$ represents the Hadamard product. $\bm{r}_k$ is the unit vector from the camera to $\bm{p}_k$. $\bm{\Phi}(\bm{r}_k)$ is a set of vectors:
\begin{align*}
& \bm{\Phi}_{d=0} = 1, \nonumber\\
& \bm{\Phi}_{d=1} = \left[\bm{\Phi}_{d=0}, -r_{k,y}, r_{k,z}, -r_{k,x}\right]^\top, \nonumber\\
&\bm{\Phi}_{d=2} =  \left[\bm{\Phi}_{d=0}, \bm{\Phi}_{d=1}, r_{k,x}r_{k,y}, r_{k,y}r_{k,z}, 2r^2_{k,z}-r^2_{k,x}-r^2_{k,y},\right. \\
& \left. r_{k,x}r_{k,z}, r^2_{k,x}-r^2_{k,y}\right]^\top, \\
& \bm{\Phi}_{d=3} = \left[\bm{\Phi}_{d=0}, \bm{\Phi}_{d=1},\Phi_{d=2},r_{k,y}(3r^2_{k,x} - r^2_{k,y}),  r_{k,x}r_{k,y}r_{k,z},\right. \\ 
& r_{k,y}(4r^2_{k,z} - r^2_{k,x}-r^2_{k,y}), r_{k,z}(2r^2_{k,z}-3r^2_{k,x}-3r^2_{k,y}), \\
& \left. r_{k,x}(4r^2_{k,z} - r^2_{k,x} - r^2_{k,y}),r_{k,z}(r^2_{k,x} - r^2_{k,y}), r_{k,x}(r^2_{k,x} - 3r^2_{k,y}) \right]^\top. 
\end{align*}
\newpage
\subsection{Projected covariance matrix}
Let $\bm{x} = [m, n]^\top$ be the image space coordinate of the pixel. $G_k = \exp^{-\frac{1}{2}(\bm{\pi}_k -  \bm{x})^\top \bm{\Sigma}^{-1}(\bm{\pi}_k -\bm{x})}$, the derivatives $\frac{\partial G_k}{\partial \bm{\Sigma}_k}$, $\frac{\partial^2 G_k}{\partial \bm{\Sigma}^2_k}$ can be written as follows:
\begin{align}
&\frac{\partial G_k}{\partial \bm{\Sigma}_k} = -\frac{1}{2}G_k \left((\bm{\pi}_k - \bm{x})^\top \frac{\partial \bm{\Sigma}^{-1}_k}{\partial \bm{\Sigma}_k} : (\bm{\pi}_k - \bm{x}) \right), \nonumber\\
&\frac{\partial^2 G_k}{\partial \bm{\Sigma}^2_k} = \frac{G^2_k}{4} \left((\bm{\pi}_k - \bm{x})^\top \frac{\partial \bm{\Sigma}^{-1}_k}{\partial \bm{\Sigma}_k} : (\bm{\pi}_k - \bm{x})  \right)^2 \nonumber\\
& - \frac{G_k}{2}\left((\bm{\pi}_k - \bm{x})^\top: \frac{\partial \bm{\Sigma}^{-1}_k}{\partial \bm{\Sigma}_k}:(\bm{\pi}_k - \bm{x})\right).
\end{align}
$\frac{\partial \bm{\Sigma}^{-1}_k}{\partial \bm{\Sigma}_k}$ is a fourth-order tensor, and $\frac{\partial^2 \bm{\Sigma}^{-1}_k}{\partial \bm{\Sigma}^2_k}$ is a sixth-order tensor. The element-wise expressions can be written as:
\begin{align*}
&\frac{\partial \left[\bm{\Sigma}^{-1}_k\right]_{i,j}}{\partial \left[\bm{\Sigma}_k\right]_{p,l}} = -\left[\bm{\Sigma}^{-1}_k\right]_{i,p}\left[\bm{\Sigma}^{-1}_k\right]_{l,j}, \nonumber\\
&\frac{\partial \left[\bm{\Sigma}^{-1}_k\right]_{i,j}}{\partial \left[\bm{\Sigma}_k\right]_{p,l}\partial \left[\bm{\Sigma}_k\right]_{g,h}} = \left[\bm{\Sigma}^{-1}_k\right]_{i,g}\left[\bm{\Sigma}^{-1}_k\right]_{h,p}\left[\bm{\Sigma}^{-1}_k\right]_{l,j} \nonumber\\ 
& + \left[\bm{\Sigma}^{-1}_k\right]_{i,p} \left[\bm{\Sigma}^{-1}_k\right]_{l,g} \left[\bm{\Sigma}^{-1}_k\right]_{h,j}.
\end{align*}

\end{appendix}
\end{document}